\UseRawInputEncoding
\pdfoutput=1

\documentclass[11pt]{article}

\usepackage[final]{acl}
\usepackage{bbm}
\usepackage[disable]{todonotes}
\newcommand*\iftodonotes{\if@todonotes@disabled\expandafter\@secondoftwo\else\expandafter\@firstoftwo\fi}  %
\makeatother

\definecolor{dandelion}{HTML}{FFD464}

\definecolor{bittersweet}{HTML}{C04F17}

\definecolor{mintgreen}{RGB}{152, 255, 152}

\usepackage{tipa}

\usepackage{times}
\usepackage{latexsym}

\usepackage[T1]{fontenc}

\usepackage[font=small,labelfont=bf]{caption}

\usepackage{tabularx}

\usepackage{framed}
\usepackage{times}
\usepackage{latexsym}
\usepackage{siunitx}
\usepackage{xspace}
\usepackage[multiple]{footmisc}
\usepackage{xcolor}
\usepackage{hyperref}

\usepackage[T1]{fontenc}

\usepackage[utf8]{inputenc}

\usepackage{microtype}

\usepackage{inconsolata}

\usepackage[noend]{algpseudocode}
\usepackage{algorithm}
\usepackage{amsfonts}
\usepackage{amsmath}
\usepackage{amssymb}
\usepackage{bm}
\usepackage{cleveref}
\usepackage{xspace}
\usepackage{amsthm}
\usepackage{mathtools}
\usepackage{relsize}
\usepackage{caption,subcaption,booktabs}
\usepackage{paralist}
\usepackage{import}
\usepackage[htt]{hyphenat}
\usepackage{booktabs}
\usepackage{graphicx}
\usepackage{tikz}
\usepackage{tabularx}
\usetikzlibrary{positioning}
\usetikzlibrary{backgrounds}
\usetikzlibrary{calc}

\usepackage{booktabs}
\usepackage{longtable}
\usepackage{multirow}
\usepackage{array}
\usepackage{ragged2e}
\usepackage{needspace}

\usepackage{enumitem}

\newcolumntype{L}[1]{>{\RaggedRight\arraybackslash}p{#1}} %
\newcolumntype{C}[1]{>{\Centering\arraybackslash}p{#1}}  %

\crefname{section}{\S}{\S\S}
\Crefname{section}{\S}{\S\S}
\crefname{table}{Tab.}{}
\crefname{figure}{Fig.}{}
\crefname{algorithm}{Algorithm}{}
\crefname{equation}{Eq.}{Eqs.}  %
\crefname{appendix}{App.}{}
\crefname{thm}{Theorem}{Theorems}
\crefname{prop}{Proposition}{Propositions}
\crefname{cor}{Corollary}{Corollaries}
\crefname{observation}{Observation}{Observations}
\crefname{assumption}{Assumption}{Assumptions}
\crefformat{section}{\S#2#1#3}

\usepackage{listings}
\usepackage[most]{tcolorbox}

\newtcolorbox[list inside=prompt,auto counter,number within=section]{prompt}[1][]{
    colbacktitle=black!60,
    fonttitle=\small,
    coltitle=white,
    fontupper=\footnotesize,
    boxsep=3pt,
    left=0pt,
    right=0pt,
    top=0pt,
    bottom=0pt,
    boxrule=1pt,
    #1,
    breakable,              %
}

\theoremstyle{definition}

\theoremstyle{definition}

\newcommand{\defn}[1]{{\textbf{#1}}}

\newcommand{\defequals}{\triangleq}

\usetikzlibrary{shapes.misc}

\definecolor{MyTawny}{HTML}{d55e00} %
\definecolor{MyGreen}{HTML}{029e73}
\definecolor{MyBlue}{HTML}{0173b2}
\definecolor{MyOrange}{HTML}{de8f05}
\definecolor{MyBronze}{HTML}{ca9161}
\definecolor{MySilver}{HTML}{949494}
\definecolor{MyRed}{HTML}{b40426}
\definecolor{MyInsignificantBlue}{HTML}{3b4cc0}

\newcommand{\answertext}[1]{{\color{MyTawny} \textit{#1}}}
\newcommand{\querytext}[1]{{\color{MyGreen} \textit{#1}}}
\newcommand{\contexttext}[1]{{\color{MyOrange} \textit{#1}}}
\newcommand{\entitytext}[1]{{\color{MyBlue}{\textit{#1}}}}

\newcommand{\context}{{\color{MyOrange}{c}}}

\newcommand{\tps}{{\text{TPS}}}
\newcommand{\basictps}{\text{BasicTPS}}
\newcommand{\dtps}{\text{Distance-based TPS}}

\newcommand{\wasserstein}{W}

\newcommand{\cost}{c}

\newcommand{\alphabet}{\Sigma}

\newcommand{\lm}{p_{\textsc{m}}}

\newcommand{\answer}{{\color{MyTawny}a}}
\newcommand{\answera}{{\color{MyTawny}a}}
\newcommand{\answerb}{{\color{MyTawny}b}}
\newcommand{\answerc}{{\color{MyTawny}c}}

\newcommand{\answerzero}{{\color{MyTawny}\texttt{0}}}

\newcommand{\answernine}{{\color{MyTawny}\texttt{9}}}

\newcommand{\query}{{\color{MyGreen}{q}}}

\newcommand{\tpscoresymb}{\rho}
\newcommand{\targetdistribution}{\mathcal{\sigma}}

\definecolor{red_fig}{HTML}{D95847}
\definecolor{blue_fig}{HTML}{5D7CE6}

\crefname{prop}{Proposition}{}

\usepackage{graphicx} %
\usepackage{amsmath, amssymb}

\usepackage[utf8]{inputenc}

\usepackage{microtype}

\usepackage{inconsolata}

\usepackage{graphicx}

\title{How Persuasive is Your Context?}

\author{
Tu Nguyen$^{*}$~\;~
Kevin Du$^{*}$~\;~ 
Alexander Miserlis Hoyle~\;~
\textbf{Ryan Cotterell}
\\
ETH Z{\"u}rich \\
\href{mailto:tunguyen1@ethz.ch}{\texttt{tunguyen1@student.ethz.ch }}~\;~
\href{mailto:kevidu@ethz.ch}{\texttt{kevin.du@inf.ethz.ch}}~\;~ \\
\href{mailto:alexander.hoyle@ai.ethz.ch}{\texttt{alexander.hoyle@ai.ethz.ch}}~\;~ 
\href{mailto:ryan.cotterell@inf.ethz.ch}{\texttt{ryan.cotterell@inf.ethz.ch}}
}

\begin{document}
\maketitle
\def\thefootnote{*}\footnotetext{These authors contributed equally to this work.}\def\thefootnote{\arabic{footnote}}
\begin{abstract}
    Two central capabilities of language models (LMs) are: (i) drawing on prior knowledge about entities, which allows them to answer queries such as \querytext{What's the official language of Austria?}, and (ii) adapting to new information provided in context, e.g., \contexttext{Pretend the official language of Austria is Tagalog.}, that is pre-pended to the question. 
In this article, we introduce targeted persuasion score ($\tps$), designed to quantify how \emph{persuasive} a given context is to an LM where persuasion is operationalized as the ability of the context to alter the LM's answer to the question.
In contrast to evaluating persuasiveness only by inspecting the greedily decoded answer under the model, $\tps$ provides a more fine-grained view of model behavior.
Based on the Wasserstein distance, $\tps$ measures how much a context shifts a model's original answer distribution toward a target distribution.
Empirically, through a series of experiments, we show that $\tps$ captures a more nuanced notion of persuasiveness than previously proposed metrics.\looseness=-1
\end{abstract}

\section{Introduction}
When answering a question, language models (LMs) often rely on knowledge acquired during pretraining \citep{petroni_language_2019, brown_language_2020, roberts_how_2020, geva_transformer_2021}. 
They are also able to adapt to user-provided context in the prompt \citep{petroni_language_2019, brown_language_2020, roberts_how_2020, geva_transformer_2021}. 
For instance, given the query \querytext{What's the official language of Austria?}, a well-pretrained LM should base the answer on its prior knowledge and respond with \answertext{German}. 
However, if the user provides additional information as a context, e.g., \contexttext{Shockingly, the most recent census shows that Tagalog is the most commonly spoken language in Austria as of 2025}, and prepends this context to the query above, the desired behavior is less clear.
On the one hand, we might expect the language model to be robust and respond \answertext{German}, reflecting the pretraining data.
On the other hand, flexibility in adapting to new knowledge may also be desired---in that case, \answertext{Tagalog} may be the preferred answer. 
Indeed, different tasks may call for different levels of context sensitivity; for example, while the desired behavior for summarizing a news article involves faithfulness to the in-context information, asking the language model to critique that news article for its objectivity may require prior knowledge.\looseness=-1

\begin{figure}
    \centering
    \includegraphics[width=\linewidth]{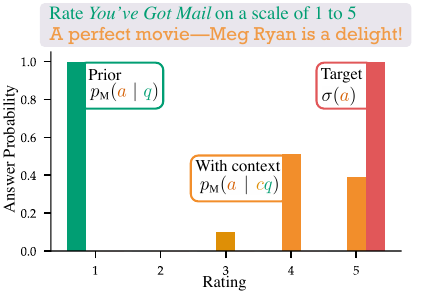}
    \vspace{-0.5cm}
    \caption{Targeted persuasion score captures the extent to which a language model can be persuaded \emph{toward} a target answer.
    With only the \querytext{query}, the LM gives the movie a 1-star rating, but if we add a \contexttext{positive review as context}, the distribution shifts toward our \answertext{target of 5}.
    $\tps$ allows the modeler to encode that \answertext{4} is closer to a \answertext{5} than a \answertext{2}.
    }
    \label{fig:tps_vs_sample}
    \vspace{-0.5cm}
\end{figure}

In both cases mentioned above, \emph{measuring} the sensitivity of a language model to a context is a fundamental problem. 
A good measure of persuasiveness should be able to capture small changes in a model's probability distribution over answers even when the greedily decoded answer remains unchanged. 
At the same time, it should also reflect not only the magnitude of the change in the model's output, i.e., how much probability mass shifts in total, but also the direction of that change, i.e., onto which answers the probability mass shifts and how similar those answers are to the original ones. 
Earlier work has approached this problem in different ways. 
Some studies investigate whether providing the context changes the greedily decoded answer under the LM  \citep[e.g.,][]{longpre_entity-based_2022, chen-etal-2022-rich, xie_adaptive_2023, zhou_context-faithful_2023, onoe_can_2023, wang_causal_2023, li_large_2022, yu_characterizing_2023}. 
While this method is intuitively straightforward, such a rigid measure misses the subtler influence of a context on the entire probability distribution over answers.
A more sensitive alternative is the persuasion score introduced by \citet{du-etal-2024-context}, which computes the KL divergence between the LM's distribution over answers before and after providing the context.
While this approach does pick up on changes in the distribution over answers even when the greedily decoded answer does not change, it does not indicate the direction of the change, i.e., whether the change moves the model closer to or further away from a specific answer.\looseness=-1

This article introduces the \defn{targeted persuasion score} ($\tps$), a metric based on Wasserstein distance that aims to achieve both the desired traits of a measure of persuasiveness mentioned above while---at the same time---being straightforward to compute.
The metric is parameterized by a user-specified cost function that encodes the relationship between different potential answers. 
For instance, the cost matrix can differentiate between numerical answers, e.g., \answertext{three} and \answertext{five}, that are two units apart on the number line.
It can also specify the relationship between different natural language answers by making use of semantic representations. 
For instance, semantically, \answertext{lovely} and \answertext{great} are closer than \answertext{lovely} and \answertext{meh}.
This flexibility allows $\tps$ to provide meaningful measures of persuasiveness across different notions of answer similarity.
Further, parameterizing TPS in a way that disregards answer similarity results in a measure of the difference in the model's target answer probability with and without the context, which is related to commonly used measures in the activation patching literature \citep{meng2023rome, wang_causal_2023, zhang2024towards, minder2025controllable}.
\looseness=-1

We establish the effectiveness of $\tps$ empirically.\footnote{\url{https://github.com/kdu4108/tps}}
The experimental section provides several case studies showcasing the strengths of $\tps$ with three different cost functions across several different question domains.
In so doing, we show that $\tps$ reveal patterns in model behavior that just analyzing the greedily decoded answers cannot.
Specifically, in one case study, we investigate how polluting a context with contradictory information can affect a model's answer to a query, and show that measuring the model's behavior with $\tps$ can reveal a ``lost-in-the-middle'' effect \cite{liu-etal-2024-lost}, i.e., an effect in which contradictory information at the start or end of a context influences a model more than information placed in the middle of a context, that is \emph{not} visible when only analyzing the greedily decoded answer. 
Taken together, these results suggest $\tps$ is a useful tool for understanding and controlling how contexts influence language model behavior.

\section{The Targeted Persuasion Score}
\label{sec:formalization}

We begin this section with the requisite mathematical background and a general definition of $\tps$, then define specific instances of the metric that are appropriate to different problem types.

\newcommand{\prefixfree}{\mathcal{P}}
\newcommand{\plm}{\overrightarrow{\lm}}
\newcommand{\Snull}{\mathcal{S}_\varnothing}
\newcommand{\tzero}{\texttt{0}}
\newcommand{\tone}{\texttt{1}}
\newcommand{\ttwo}{\texttt{2}}
\newcommand{\tthree}{\texttt{3}}
\newcommand{\tfour}{\texttt{4}}
\newcommand{\tfive}{\texttt{5}}
\newcommand{\tsix}{\texttt{6}}
\newcommand{\tseven}{\texttt{7}}
\newcommand{\teight}{\texttt{8}}
\newcommand{\tnine}{\texttt{9}}

\subsection{Preliminaries and \tps}
\paragraph{Language Modeling.}
Let $\alphabet$ be an \defn{alphabet}, a finite, non-empty set of symbols.
A \defn{language model} $\lm$ is a probability distribution over $\alphabet^*$, the set of all strings with symbols drawn from $\alphabet$.
We denote the \defn{prefix probability} of a language model $\lm$ as $\plm(x) \defequals \sum_{y \in \alphabet^*} \lm(xy)$.
Further, we write $\plm(x \mid y) \defequals \frac{\plm(yx)}{\plm(y)}$ for the conditional prefix probability when $\plm(y) > 0$. 
For strings $x, y \in \alphabet^*$, we write $x \preceq y$ if $x$ is a prefix of $y$, i.e., that $y$ begins with $x$.
Finally, a set of strings $\prefixfree \subseteq \alphabet^*$ is called a \defn{prefix-free cover} \citep{vieira2025language} if the following two properties hold:\looseness=-1

\begin{itemize}
\item \textbf{Prefix-free}: $x, y \in \prefixfree \implies x \not\preceq y \wedge y \not \preceq x$,
\item  \textbf{Cover}: For all $x \in \alphabet^*$, there exists a $y \in \prefixfree$ such that $y \preceq x$.
\end{itemize}
Observe that, for a language model $\plm$, we have $\sum_{x \in \prefixfree} \plm(x \mid y) = 1$ if $\prefixfree$ is a prefix-free cover.

\paragraph{Wasserstein Distance.}
We introduce the notion of the Wasserstein distance in the abstract \citep{Kantorovitch1958OnTT,vaserstein1969markov}.
Let $q_1$ and $q_2$ be discrete probability distributions over a set $\mathcal{X}$.\footnote{The support of $q_1$ and $q_2$ may be distinct.}
Let $\Pi(q_1, q_2)$ be the set of all marginal-preserving joint probability distributions over $\mathcal{X} \times \mathcal{X}$, i.e., the set of all functions $\gamma \colon \mathcal{X} \times \mathcal{X} \rightarrow [0, 1]$ such that 
\begin{equation}
\begin{aligned}
\sum_{y \in \mathcal{X}} \gamma(x, y) &= q_1(x) \\
\sum_{x \in \mathcal{X}} \gamma(x, y) &= q_2(y) \\
\gamma(x, y) &\geq 0, \quad \forall x, y \in \mathcal{X} \times \mathcal{X}.
\end{aligned}
\end{equation}
Then, we define the \defn{Wasserstein distance} as 
\begin{equation}
\wasserstein(q_1, q_2) \defequals \inf_{\gamma \in \Pi(q_1, q_2)}  \sum_{(x, y) \in \mathcal{X} \times \mathcal{X}}  \gamma(x, y) \cost(x, y),
\end{equation}
where $\cost(x, y) \colon \mathcal{X} \times \mathcal{X} \rightarrow \mathbb{R}_{\geq 0}$ is a non-negative \defn{cost function}.
Chosen by the modeler, $\cost(x, y)$ specifies the work needed to move probability mass from an outcome $x$ in the support of one probability distribution to an outcome $y$ in the other.\footnote{When $\mathcal{X}$ is finite, it can be helpful to think of the cost function as a $|\mathcal{X}|\times|\mathcal{X}|$ matrix.} An element $\gamma \in \Pi(q_1, q_2)$ is often called a transportation plan where $\gamma(x, y)$ indicates the amount of probability mass moved from $q_1(x)$ to $q_2(y)$. Loosely, the distance is therefore the most efficient way of shifting mass between the two distributions under $\cost(x, y)$.\looseness=-1

\paragraph{Targeted Persuasion Score.}
We now introduce the targeted persuasion score in the context of language modeling. 
With $\query \in \alphabet^*$, we denote a query string, e.g., \querytext{What is the capital of Florida?}, and, with $\context \in \alphabet^*$, we denote a context string, e.g., \contexttext{Breaking News: An indebted Florida has forged a deal with Disney and changed its capital to The Magic Kingdom}.
The concatenation of both the {\color{MyOrange}context} and {\color{MyGreen}query} is denoted by a bare juxtaposition $\context\query$.
Finally, with $\answer \in \alphabet^*$, we denote an {\color{MyTawny}answer} string.
Let $\lm$ be a language model over alphabet $\alphabet$ and let $\prefixfree \subset \alphabet^*$ be a finite, prefix-free cover.
Thus, both  $\plm(\cdot \mid \query)$ and $\plm(\cdot \mid \context\query)$ are distributions over $\prefixfree$.\footnote{In considering $\plm$ as distributions over a prefix-free cover $\prefixfree$, this formulation is agnostic to tokens boundaries and, thus, applies to multi-token answers as well. 
To compute the probability of a multi-token answer, for a given question, one first enumerates a set of possible distinguished answers and construct a prefix-free cover that contains that set. Then, using the algorithm described in \citet{vieira2025language}, one can compute the probability of each string in the prefix-free cover. $\tps$ can thus be computed over answers regardless of the number of tokens per string.}
We call $\plm(\cdot \mid \query)$ the \defn{prior} probability distribution and $\plm(\cdot \mid \context\query)$ \defn{context-conditional} probability distribution.
Let $\targetdistribution$ be a user-specified \defn{target} distribution over $\prefixfree$.
Then, we define the \defn{targeted persuasion score} ($\tps$) as 
\begin{equation}
\begin{aligned}
\tpscoresymb(\context, &\query, \plm, \targetdistribution) \\
&\defequals
\wasserstein(\plm(\cdot \mid \query), \targetdistribution) - \wasserstein(\plm(\cdot \mid \context\query), \targetdistribution).
\end{aligned}\label{eq:tps}
\end{equation}
Imagine the target distribution as a location. The left hand-side of eq. \ref{eq:tps} measures how far the LM is from that point \emph{without} a context, and the right hand side the distance from the point \emph{with} a context.

\subsection{BasicTPS}
Specific instances of $\tps$ depend on the choice of the target distribution $\targetdistribution$ and the cost function $\cost$.
We first consider the case where there exists a distinguished answer $\answer^\star \in \prefixfree$ such that $\targetdistribution(\answer^\star) = 1$ and the cost function is defined as 
\begin{align}
\cost(\answerb, \answerc) = \begin{cases}
    1 & \text{if} \quad \answerb = \answer^\star, \answerc \neq \answer^\star \\
    0 & \text{otherwise}.
\end{cases}
\end{align}
In words, this cost function assigns \begin{inparaenum}[(i)] 
\item a cost of 1 to move probability mass from a non-target answer to the target answer, and
\item no cost associated with moving probability mass in other cases.
\end{inparaenum} 
This version of $\tps$ takes values in $[-1,1]$ and is called $\basictps$.
A score of $1$ means that the context has shifted the model's prior probability distribution with all mass on a non-target answer to a context-conditional probability distribution with all mass on the target answer. Conversely, a score of $-1$ means the context has shifted the model's prior probability distribution with all mass on the target answer to a context-conditional probability distribution with all mass on a non-target answer.

\begin{figure*}
    \begin{subfigure}[t]{0.32\textwidth}
        \includegraphics[width=\textwidth]{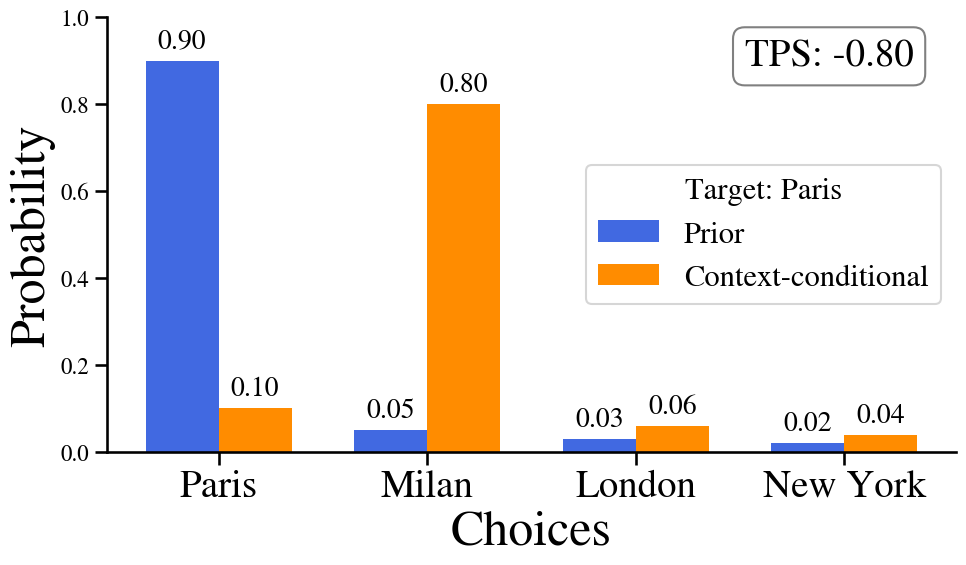}
        \caption{If the target is \answertext{Paris} and the prior has high probability on \answertext{Paris}, yet a context moves the mass away from the target, then $\basictps$ will be highly negative.}
        \label{fig:city_1}
    \end{subfigure}
    \hfill
    \begin{subfigure}[t]{0.32\textwidth}
    \includegraphics[width=\textwidth]{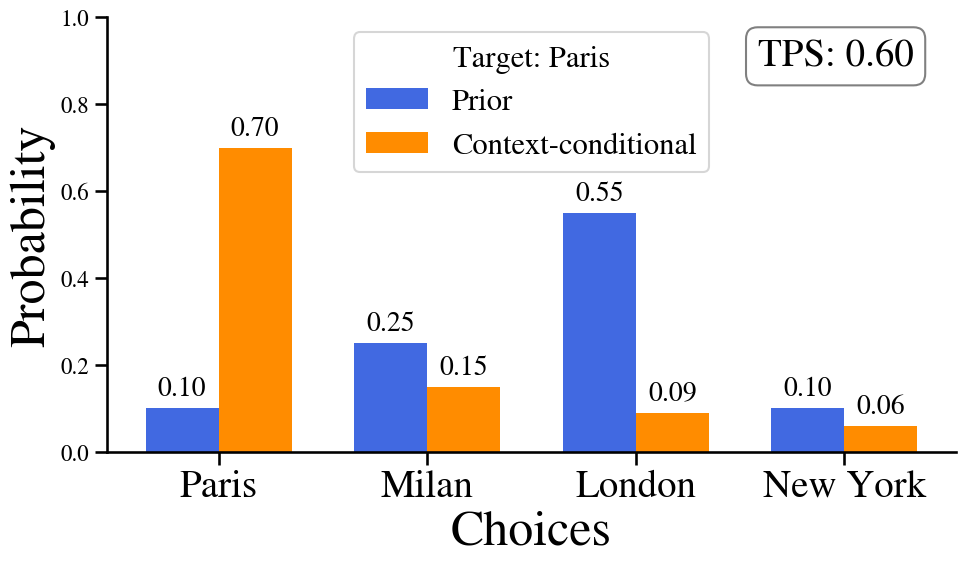}
        \caption{If the target is \answertext{Paris},  the prior has low probability on \answertext{Paris}, and a context adds probability mass to the target, then the $\basictps$ will be highly positive.}
        \label{fig:city_2}
    \end{subfigure}
    \hfill
    \begin{subfigure}[t]{0.32\textwidth}
    \includegraphics[width=\textwidth]{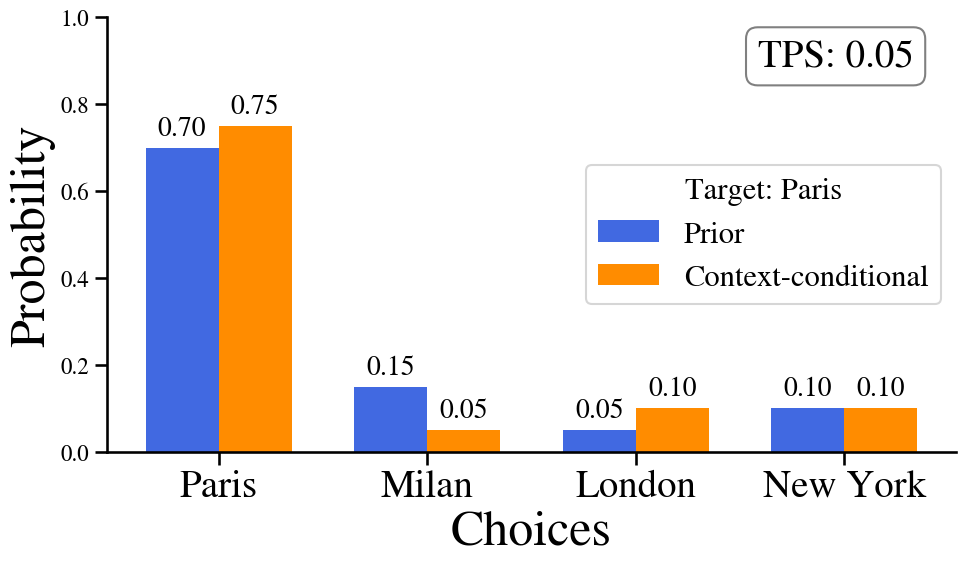}
         \caption{If the target is \answertext{Paris}, the prior has high probability on the target, and a context adds little mass to the target, then the $\basictps$ will be slightly positive.}
        \label{fig:city_3}
    \end{subfigure}
    \caption{Examples of $\tps$ for 3 different scenarios of prior and context-conditional probability distributions.}
    \label{fig:basictps_scenarios}
\end{figure*}

\begin{figure}
    \centering
    \includegraphics[width=\linewidth]{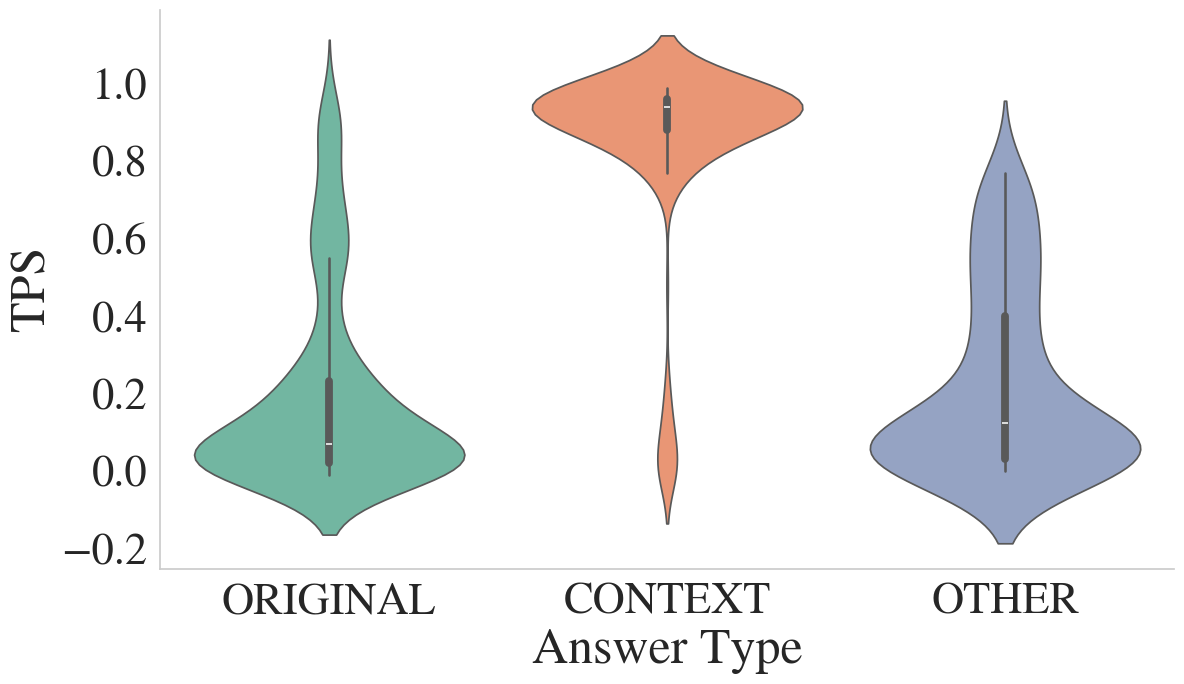}
    \vspace{-0.7cm}
    \caption{
     $\basictps$ is \emph{high} when the model's sampled answer agrees with the context (center), and \emph{low} otherwise (left, right).
    }
    \label{fig:tps_vs_sample}
\end{figure}

\subsection{Distance-based TPS}\label{sec:ordinal-tps}
In some tasks, the relationship between answers is more complex.
For instance, in the context of assessing machine translation, some possible translations are better than others. 
In these cases, incorrect answers are not equally wrong.
The cost function applied in $\basictps$, however, models the case where there exists a single, correct answer.\looseness=-1

\paragraph{Ordinal relationships.} 
Next, we consider answers that encode ordinal relationships.
First, define a scale set $\mathcal{S} \subset \alphabet^*$ and an out-of-scale sentinel $\varnothing$, with $\Snull \defequals \mathcal{S}\cup\{\varnothing\}$. Then define a map $s \colon \alphabet^* \rightarrow \mathcal{S}\cup\{\varnothing\}$, with $s(x) = x$ if $x \in \mathcal{S}$ and $\varnothing$ otherwise.
Suppose we have a distance function $d(\cdot, \cdot) \colon \Snull \times \Snull \rightarrow \mathbb{R}_{\geq 0}$.
We define a cost function as $c(a,b) = d(s(a),s(b))$.\footnote{We equivalently change the support of the prior and context-conditional distributions to $\Snull$, e.g., $\plm(s(\cdot) \mid \context)$} %
One special case of the above that we will make use of in this paper is based on ordinal numbers, where the answer space consists of integers on a scale, i.e., $\mathcal{S} = \{\answerzero,\ldots,\answernine\}$.\footnote{Fixed-width \texttt{\textbackslash texttt} numerals are strings in $\alphabet$; those in standard font are elements of $\mathbb{Z}$.} 
To motivate the ordinal scale, we investigate the behavior of an LM in rating a movie.
Consider a query \querytext{On a scale of \tzero{} to \tnine, what is the rating of Titanic} and a context \contexttext{This movie was a pretty good movie overall}. Here, $\Snull$ is equipped with a total order, and the distance between pairs of elements reflects how far apart they are on this ordinal scale.
For instance, if the target rating is \answertext{\tnine}, then an answer of \answertext{\tseven{}} should be considered closer than an answer of \answertext{\tthree}. Given a function $n \colon \mathcal{S} \rightarrow \{0,\ldots,9\}$ that maps strings to integers, we define the distance between two ratings $\answera, \answerb \in \mathcal{S}$ as the normalized absolute difference of their corresponding integer representations, i.e., $d(\answera,\answerb) = \frac{1}{9}\lvert n(\answera) - n(\answerb) \rvert$. Then the cost of moving one unit of mass between two answers is the distance normalized by the maximum distance between two ratings (0 and 9), i.e.,
\begin{equation}
\cost(\answera,\answerb) =
\begin{cases}
\frac{\lvert n(\answera) - n(\answerb) \rvert}{9} &\text{if } \answera, \answerb \in \mathcal{S}\label{eq:ordinaltps} \\
0& \text{otherwise}.
\end{cases}
\end{equation}

\paragraph{Semantic relationships.} 
In general, beyond the specific case of ordinal-valued answers discussed in \Cref{sec:ordinal-tps}, it may be difficult to devise a distance $d$ to construct the cost
function.
One solution is to use word representations \citep{wmd-kusnerb15}, i.e.,\looseness=-1
\begin{equation}\label{eq:semantic-cost}
\cost(\answera, \answerb) = 1 - \frac{\mathbf{e}(\answera) \cdot \mathbf{e}(\answerb)}{||\mathbf{e}(\answera)||_2 ||\mathbf{e}(\answerb)||_2},
\end{equation}
where $\mathbf{e} \colon \alphabet^* \rightarrow \mathbb{R}^D$ maps strings real-valued representations, e.g., with the method given in \citet{reimers-2019-sentence-bert}. 
Using \Cref{eq:semantic-cost} with well chosen representations means that persuading the model to shift among semantically similar answers is easier than persuading the model to shift among answers which are further away semantically.\looseness=-1

\section{Two Toy Case Studies }
In this section, we build intuition for $\tps$ with two controlled toy experiments that explore two different choices of cost function $\cost$. 

\subsection{Case Study \#1}

\paragraph{Setup.} 
In our first case study, we consider 500 queries from the \querytext{officialLanguage} relation of the YAGO knowledge graph \cite{yago2007}, as extracted in \citet{du-etal-2024-context}.
For each query, we prepend a context $\context$ which disagrees with the ground truth answer to the query $\query$.
We then prompt the Qwen-2.5 7B Instruct model \citep{qwen2.5} with $\context\query$ in addition to necessary instructions for this model.
We take $\prefixfree$ to be the set of all tokens in the alphabet of the model, Qwen-2.5 7B for this experiment.
We compute the greedily decoded answer under the model and track whether the answer is the one suggested by the context, the one originally preferred by the model, or something else.
For example, if we prompt a model with: \contexttext{The official language of Brazil is French.} \querytext{What is the official language of Brazil?}, we then evaluate whether the model's outputted answer matches \answertext{French}, as suggested by the context, \answertext{Portuguese}, the model's preferred answer without context, or something else. 
We also compute the $\basictps$ where the target distribution $\targetdistribution$ is the probability distribution that places probability 1 on \answertext{French}.
As a sanity check, we expect that $\basictps$ scores are higher for contexts that successfully change the greedily decoded answer under the model to match the context and lower for contexts that do not.

\paragraph{Results.}
From \Cref{fig:tps_vs_sample}, we see that the contexts where the greedily decoded answer changes to agree with the context generally have higher $\basictps$ scores than the case where it does not change, or changes to not agree. 
This is consistent with our expectation and supports the validity of the $\basictps$ as a measure for a context's persuasiveness.
Further, \Cref{fig:tps_vs_sample} shows the existence of examples where the model's greedily decoded answer did not agree with the context, yet have high $\tps$ scores. 
This suggests that judging persuasiveness by greedy decoding alone may obscure an effect seen by the $\basictps$, namely that some contexts may substantially change the model's probability of the correct answer without making it the top outcome.\looseness=-1

\subsection{Case Study \#2}\label{sec:word-sence-case-study}

In our next case study, we consider a task based on word-sense disambiguation \cite{navigli2009wsd}.
Our goal is to measure the extent to which placing a word in a sentence disambiguates its word sense.
We proceed with a simple example. 
Consider the word \entitytext{run} from WordNet \citep{miller-1992-wordnet}, which annotates words with multiple senses and corresponding example usages.
The four senses of the word \entitytext{run} in WordNet are as follows:
\begin{description}[labelwidth=\widthof{(sense \answertext{direct})},
                    leftmargin=!,
                    labelsep=0.75em,
                    itemsep=-0.15em]
                    
  \item[(sense \answertext{move})]   \answertext{move fast by using one’s feet}
  \item[(sense \answertext{direct})] \answertext{direct or control a business or activity}
  \item[(sense \answertext{score})]  \answertext{a score in baseball made by a runner reaching home base}
  \item[(sense \answertext{trip})]   \answertext{a short trip or errand}
\end{description}

\noindent Senses are then turned into the following query:\looseness=-1
\begin{quote}
\querytext{Choose among 4 definitions X, Y, Z, T. 
Definition X: move fast by using one's feet. 
Definition Y: direct or control a business or activity. 
Definition Z: a score in baseball made by a runner reaching home base.
Definition T: a short trip or errand. 
The suitable definition of \entitytext{run} is?}
\end{quote}
As the context, we consider sentences that unambiguously invoke one of the senses of the word \entitytext{run}:
\begin{description}[labelwidth=\widthof{(sense \answertext{direct})},
                    leftmargin=!,
                    labelsep=0.75em,
                    itemsep=-0.15em]
    \item[(sense \answertext{move})]   \contexttext{Taylor went for a run in a hurry around the forest trail for training.}
    \item[(sense \answertext{direct})] \contexttext{She runs a successful company with over twenty employees.}
    \item[(sense \answertext{score})]  \contexttext{The batter scored a run in the fifth inning.}
    \item[(sense \answertext{trip})]   \contexttext{I just need to make a quick run to the pharmacy before dinner.}
\end{description}
During experimentation, we automatically generate 100 such sentences for each sense using GPT-5 \cite{OpenAI2025GPT5}
We first run a sanity check to verify that in this setting of word sense disambiguation, $\tps$ should be positive for the target sense and negative for non-target senses. 
We collect 100 words with four possible senses from WordNet \cite{miller-1992-wordnet}. 
We use a one-sided $t$-test to measure that the $\tps$ toward the \emph{target} sense is greater than 0 and that the $\tps$ toward the \emph{non-target} senses is less than zero, i.e., $p < 0.01$ with a Bonferroni correction for the four instantiations across the 100 senses.
Across the 400 senses, the $\tps$ for the target sense is significantly greater than 0 in all but 7.75\% of the cases, and $\tps$ for the non-target senses are always significantly less than zero---the full word list and positive $\tps$ scores are in \cref{app:word-sense}.

\begin{figure}[t]
    \centering
    \includegraphics[width=\linewidth]{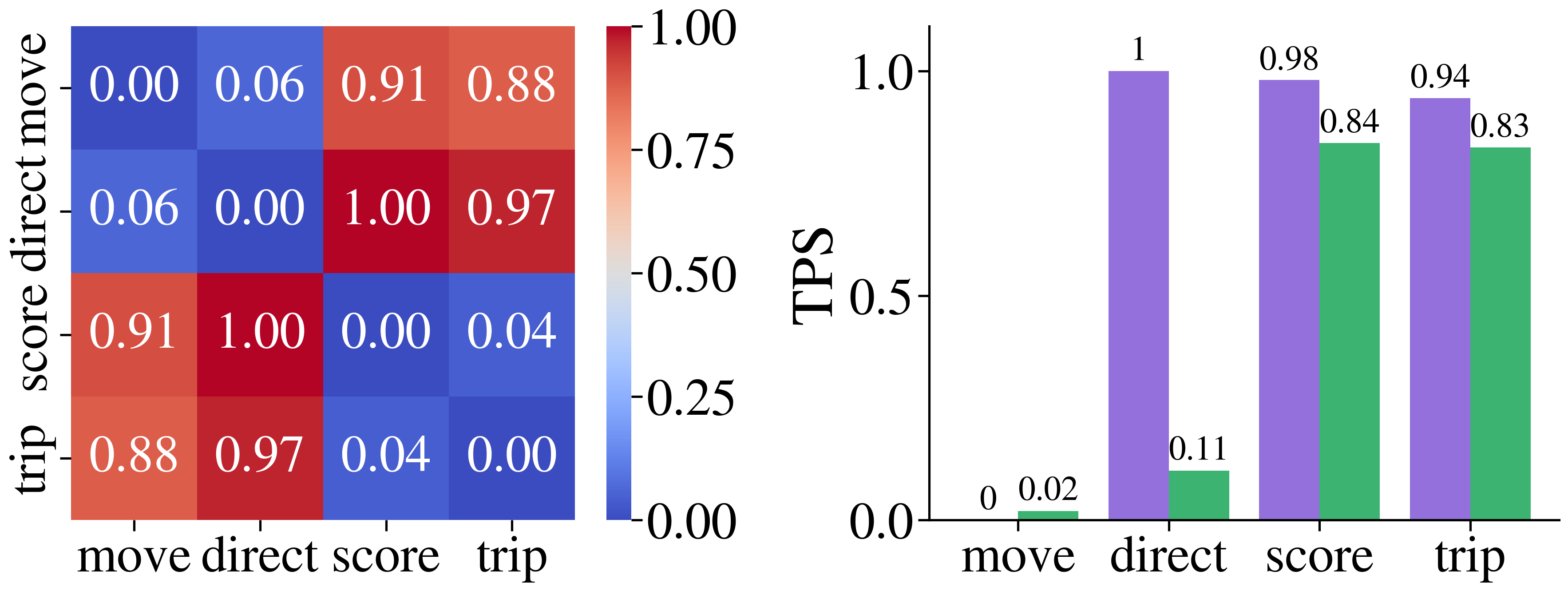}
    \vspace{-0.8cm}
    \caption{The heatmap (left) shows the pairwise cosine distance between pairs of senses for the word \entitytext{run}. The bar plot (right) shows the \textcolor[HTML]{9B59B6}{$\basictps$} (purple) and \textcolor[HTML]{27AE60}{$\dtps$} (green) for 4 groups of contexts, each group contains 100 sentences aiming to persuade the model toward a sense. We report the average $\dtps$ across 100 contexts of each sense, with each $\dtps$ computed using the cost function in \Cref{eq:semantic-cost}.}
    \label{fig:word_sense}
\end{figure}

Next, we compare $\basictps$ and $\dtps$ across the four senses.
\Cref{fig:word_sense} demonstrates for the word \entitytext{run} (which has a prior distribution concentrating $93\%$ of its probability mass on the sense \answertext{move}) both the cosine distances between the word sense pairs, as well as $\tps$ for different target senses. 
That is, the left panel shows the cost function encoding the pairwise semantic distances between senses, per a Qwen-based sentence embedding model \cite{zhang2025qwen3embeddingadvancingtext},\footnote{\url{https://huggingface.co/AlexWortega/qwen3k}} and the right panel plots the two types of $\tps$ values for each of the four example sentences. 
As seen in the figure, the $\basictps$ does not discriminate between how similar the senses are to each other and thus produces high values for each of the senses (which did not already have a high concentration of probability mass).
However, since the $\dtps$ encodes cosine distance into its cost function, senses that are more similar to the answer with highest probability mass in the prior distribution (\answertext{move}) have lower $\dtps$ and vice versa.
Indeed, we compute the cosine distance between word senses for $100$ words, as well as the $\dtps$ and $\basictps$ and find a Spearman correlation of $-0.96$ between the cosine distance of two word senses and the difference in $\basictps$ and $\dtps$. 
This suggests that, as expected, the $\dtps$ will account for the similarity between semantic embedding similarity of word senses while $\basictps$ will not.

\section{A More Realistic Study}%
\label{sec:otps_experiments}
Now, we illustrate that $\tps$ can be a useful tool for the analysis of language model behavior in a more realistic setting. 
More specifically, we run a more controlled study to analyze how different contexts influence Qwen-2.5 models.
As our testbed, we consider a movie rating task.
We consider 1,000 randomly selected movies from the IMDb movie review dataset \citep{imdb} and synthetically generated movie reviews using ChatGPT \citep{openai_gpt4_2023}.
This setting lets us study the relationship between attributes of reviews in the context, e.g., the sentiment of the review as well as the model's answer to a question where an ordinal answer is expected, i.e., \querytext{Rate \entitytext {\{movie\}} from \tzero{} to \tnine}. In this experiment, the relative distances among answers in the answer space (\answerzero{} to \answertext{\tnine}) are captured by the ordinal cost function (Eq.~\ref{eq:ordinaltps}).%

\subsection{Persuasiveness and Number of Examples}
\label{sec:exp_tps_vs_k}
\paragraph{Setup.} A natural question to ask is the following: how much does increasing the number of in-context examples drive the model toward a target answer?
Using the movie review dataset, we investigate whether $\tps$ (with the target answer set to \answernine, the highest rating) increases as the number of positive reviews grows.
Building on the result from the previous section---that a single negative in-context review tends to be more persuasive than a single positive one---we also ask whether this effect strengthens, diminishes, or remains unchanged when the context consists of multiple reviews.
We divide movies from the IMDb dataset into above-average and below-average sets, based on whether the model's prior distribution over the possible ratings (\answerzero{} to \answernine{}) has an expected value above or below $4.5$.
For $k \in \{1, \ldots, 10\}$, we randomly sample $k$ negative (resp.\ positive) synthetic reviews for each movie with a high (resp.\ low) prior rating.
We also construct deliberately \emph{high-variance} contexts by sampling $k \in \{4, \ldots, 10\}$ reviews.
In these high-variance contexts, two thirds of the reviews share the designated polarity (positive or negative), while one third have the opposite polarity.
Finally, we evaluate how noisy negative (resp.\ positive) reviews persuade the model to decrease (resp.\ increase) a previously high (resp.\ low) rating, using the $\dtps$ with a target answer of \answerzero{} (or \answernine).
Here, we use Qwen-2.5 7B Instruct.\looseness=-1

\begin{figure}
    \centering
    \includegraphics[width=\linewidth]{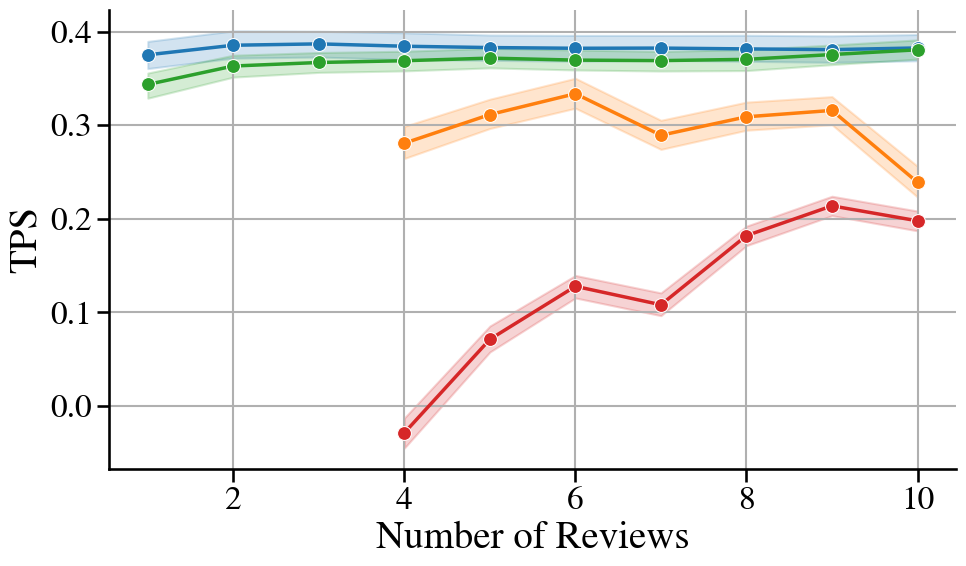}
    \caption{How $\dtps$ varies with number of reviews distinguished by four types of reviews: \textcolor[HTML]{1f77b4}{negative}, \textcolor[HTML]{2ca02c}{positive}, \textcolor[HTML]{ff7f0e}{noisy negative}, and \textcolor[HTML]{d62728}{noisy positive}.}
    \label{fig:otps_vs_k.png}
\end{figure}

\paragraph{Results.}
When $k$ is small, the uniformly \emph{negative} reviews have higher $\dtps$ scores than uniformly \emph{positive} reviews (\cref{fig:otps_vs_k.png}), suggesting that small numbers of negative contexts are more persuasive than small numbers of positive contexts.
In addition, increasing the number of positive reviews in a uniformly positive context can weakly improve persuasiveness for this model and dataset.
As a result, the effect of negative contexts having higher $\dtps$ than positive contexts disappears as the number of reviews increases: the $\dtps$ for positive contexts increases weakly as $k$ increases, but stays mostly the same regardless of the number of negative reviews in a uniformly negative context.   
We further observe the effect of noise in a context.
As expected, uniform contexts have higher $\dtps$ than the noisy ones.
Interestingly, for low $k$, when the context is mostly positive but features some negative reviews (and is directed toward a high target rating of \answernine{}), the $\dtps$ is much lower than when the context is mostly negative but features some positive reviews (and directed toward a low target rating of \answerzero{}). 
This further suggests that negative information in-context may be more persuasive than positive information in-context (for Qwen 2.5 7B Instruct).
Like with the non-noisy contexts, however, this effect diminishes as $k$ increases.
In \Cref{app:moremodels}, we show similar findings for additional model families and sizes (Llama-3.2 3B and Gemma 7B).\looseness=-1

\subsection{Concatenated vs. Individual Reviews}\label{sec:concat_vs_individual}
\paragraph{Setup.} 
Next, we examine how the individual reviews that are concatenated together to form a context might influence the model's behavior. 
To this end, we explore how the $\tps$ of a single context consisting of $K$ concatenated reviews compares to the mean $\tps$ of each of the $K$ component reviews separately. 
If the $\tps$ of the concatenated reviews does not equal the mean $\tps$ of the individual reviews, this suggests that not all reviews are equally influential in the concatenated context.
For each of the 1000 movies in our experiment, we construct contexts consisting of between 4 and 10 concatenated reviews.
Like in \Cref{sec:exp_tps_vs_k}, we construct four groups of contexts: uniformly positive, uniformly negative, noisy positive, which contain up to $K/3$ negative reviews, and noisy negative, which contain up to $K/3$ positive reviews.
Contexts in each group are created by randomly sampling the desired number of positive and negative reviews.\footnote{
As before, contradictory reviews are ordered in the middle of the concatenated contexts.}
We then compute the $\dtps$ of Qwen 2.5 7B Instruct for these concatenated reviews and the mean of the $\dtps$ for each individual review.\looseness=-1

\begin{figure}
    \centering
    \includegraphics[width=\linewidth]{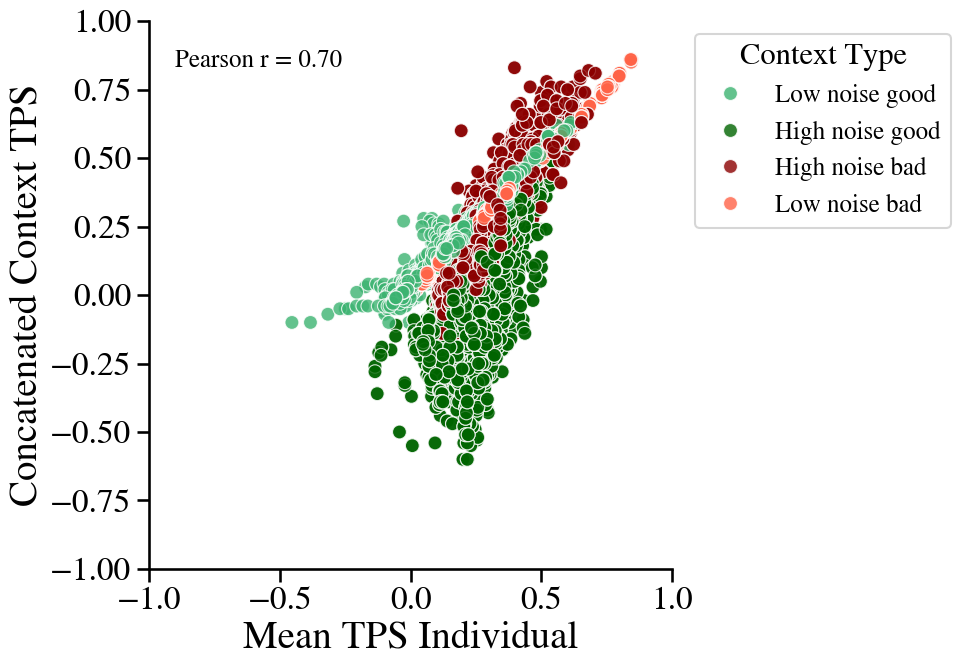}
    \vspace{-0.7cm}
    \caption{The $x$-axis is the mean $\tps$ of individual reviews and the $y$-axis is the $\tps$ of the concatenated review with of the individual reviews. Notably, the slope of a regression through the high-noise points is greater than 1, suggesting that noisy reviews which are influential to the $\tps$ on their own become less so when part of a concatenation of multiple reviews.\looseness=-1
    }
    \label{fig:Correlation_mean_TPS_vs_concat_TPS}
\end{figure}

\paragraph{Results.} 
For the purely positive and purely negative contexts, the $\dtps$ of concatenated reviews is approximately the same as the mean $\dtps$ of its individual sub-reviews (\Cref{fig:Correlation_mean_TPS_vs_concat_TPS}).
This is largely expected; if the reviews are roughly similar in positivity (or negativity) level, the model ought to give similar scores to the individual reviews and to the concatenated review.
That said, the noisy contexts still show a strong, monotonic relationship between the $\tps$ of the concatenated reviews and the mean $\tps$ of its individual reviews. 
However, the relationship between these two quantities is not the identity.
That the slope is greater than 1 means that while the contradictory reviews drive down the mean $\tps$ of the individual reviews, the $\tps$ of the concatenated review does not diminish as much.
This suggests that the influence of contradictory reviews on $\tps$ is less when in a concatenated context than when on their own.\looseness=-1

\subsection{A Lost-in-the-middle Effect}
\label{sec:lost_in_the_middle}

\paragraph{Motivation.} 
In our final experiment, we aim to identify a pattern in model behavior which analyzing the greedily decoded answer alone would not be able to find.
Inspired by the lost-in-the-middle effect identified in retrieval-augmented generation \citep{liu-etal-2024-lost},\footnote{Important information is ignored when placed in the middle of a long context.} we investigate whether a similar effect might also exist in our setting.
Specifically, we ask whether placing a contradictory review at the beginning or end of a context influences the model's answer significantly more than placing it in the middle.

\paragraph{Setup.} 
To test for a lost-in-the-middle effect, for each movie, we construct a set of 10 synthetic reviews where 9 are positive and 1 is negative.
From each set of 10 reviews, we construct 10 permutations where the order of the 9 good reviews is fixed across the permutations and each permutation features the negative review at a different position. 
We then compute the $\tps$ and the greedily decoded rating for each permutation. 
Our goal is to find permutations with outlier scores. 
To this end, we use a standard anomaly detection method:
we compute the median absolute deviation \cite[$\mathrm{MAD}$;][]{leys2013mad} for both the $\tps$ and decoded rating across the 10 permutations and flag the permutations with a $\tps$ (or a change in the LM's rating of the movie) more than $3\cdot\mathrm{MAD}$ away from the median \citep{Hoaglin2013Volume1H}.
If a permutation with the negative review at position $i$ is detected as an outlier compared to the permutations with the negative review at other positions, then this suggests that putting the negative review at position $i$ is especially persuasive compared to the other positions.
We then identify which permutations have been flagged according to the $\dtps$, $\basictps$, and decoded rating. 
We compare the positions of the bad reviews which most elicit outlier flags for the $\dtps$ and the decoded rating.\looseness=-1

\begin{figure}
    \centering
    \includegraphics[width=\linewidth]{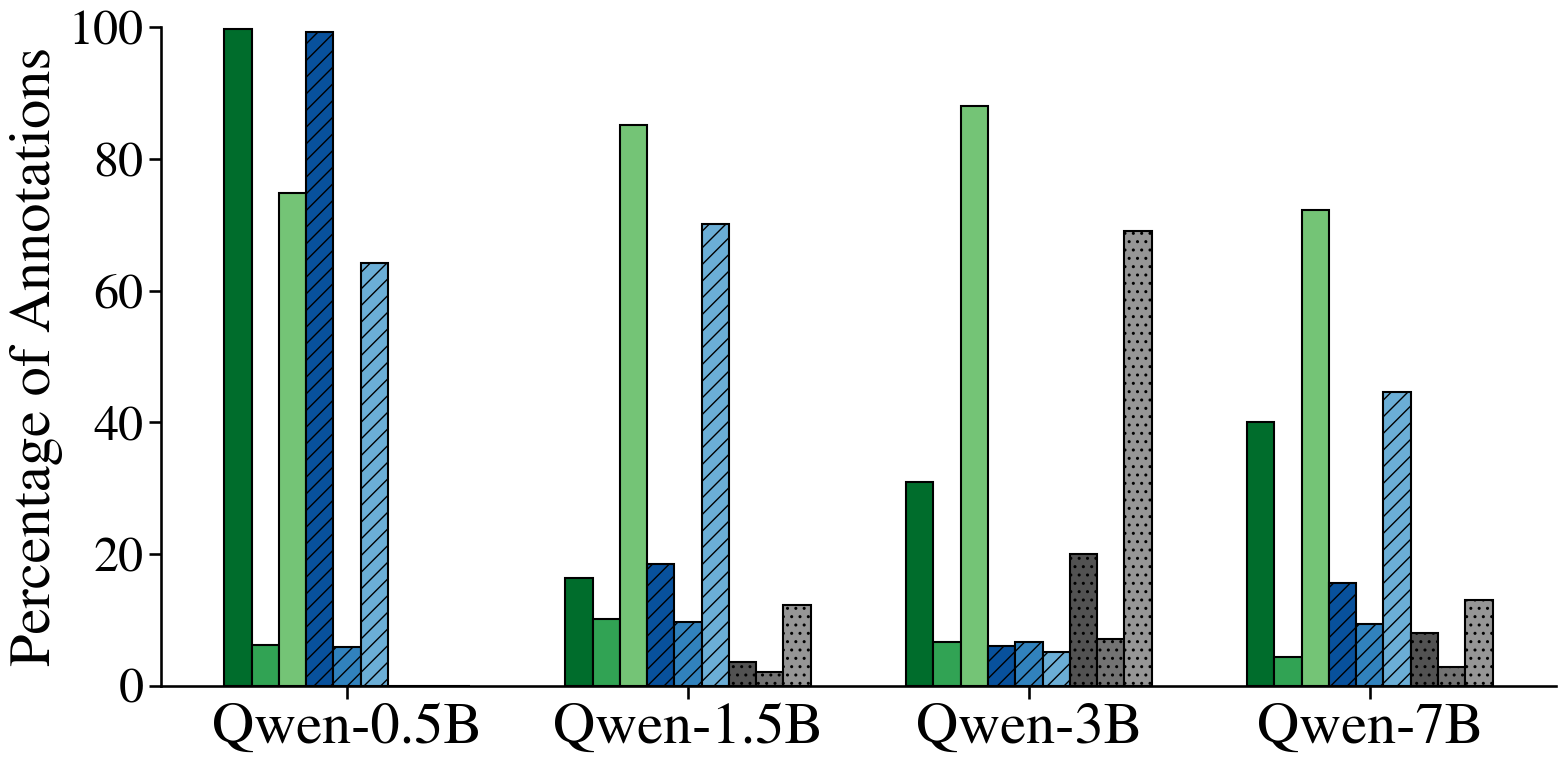}
    \caption{Percentage of anomalies detected across Qwen model sizes, split by metric (decoded rating, $\basictps$, and $\dtps$) and by position of the contradictory review: \textcolor[HTML]{525252}{Rating (first)}, \textcolor[HTML]{737373}{Rating (middle)}, \textcolor[HTML]{969696}{Rating (last)}; \textcolor[HTML]{08519c}{$\basictps$ (first)}, \textcolor[HTML]{3182bd}{$\basictps$ (middle)}, \textcolor[HTML]{6baed6}{$\basictps$ (last)}; \textcolor[HTML]{006d2c}{$\dtps$ (first)}, \textcolor[HTML]{31a354}{$\dtps$ (middle)}, \textcolor[HTML]{74c476}{$\dtps$ (last)}. 
    This plot indicates that, across four model sizes ($x$-axis), a significant percentage of the dataset ($y$-axis) consists of anomalies captured by the $\dtps$ and $\basictps$ but not by the decoded ratings increase.
    Further, this plot shows that through $\dtps$, one can see a ``lost-in-the-middle'' effect where contradictory contexts inserted at the start and end of a context influence $\dtps$ more, while such a pattern is invisible when looking at decoding-only measures and less clear with $\basictps$.}
    \label{fig:Anomaly}
\end{figure}

\paragraph{Results.}
With $\dtps$, there is a clear ``lost-in-the-middle'' effect with the contradictory context (\Cref{fig:Anomaly}).
Specifically, across all four model sizes, when the contradictory review is in the first or last position of the ten reviews, the $\dtps$ is significantly lower than when the contradictory review is in the middle positions.
This is especially prominent in the smallest model (Qwen-2.5 0.5B Instruct).
Notably, such a finding could not be observed from greedily decoding the movie ratings.
Beyond demonstrating that a ``lost-in-the-middle'' effect with the contradictory context exists in this movie review setting, this finding illustrates how the $\dtps$ can help identify patterns in model behavior that would otherwise remain concealed when measuring model behavior with decoding alone.

\section{$\tps$ in an Applied Setting}\label{sec:section-five}

As the instruction-following capabilities of LLMs have improved, they have been increasingly used for \emph{automated text annotation}, a basic task in the computational social sciences \cite[][\emph{inter-alia}.]{rytting2023codingsocialsciencedatasets,pangakis2023automatedannotationgenerativeai,ollion2023chatgpt,ziems-etal-2024-large,tan-etal-2024-large,halterman2025codebookllmsevaluatingllms,McLachlan2025annotator,baumann2025largelanguagemodelhacking}.\footnote{While individual examples abound, we have elected to cite works that take a broader and more measured view of this practice.}
For example, an LLM might code a bill in the U.S. congress as relating to a specific topic, like the \textsc{macroeconomy} \cite{egami-etal-2023-imperfect}.\looseness=-1

A common feature of such work is to adapt detailed \emph{codebooks} that describe the annotation task---originally designed for human annotators---into prompts for language models.
Such codebooks will often provide technical definitions of concepts, like \textsc{left-wing} or \textsc{populist}, that are specific to the given context and domain (and which may differ from lay understanding). 
Here, the (implicit) belief is that such instructions will encourage the model to align with experts' characterizations of text.
However, it remains unclear whether models faithfully follow such instructions \cite{Turpin2023LanguageMDA,kung-peng-2023-models}, with behavior even varying over semantically-similar prompts  \cite{sclar2024quantifyinglanguagemodelssensitivity,Abraham2025,Atreja_2025,barrie2025promptstabilityscoringtext,baumann2025largelanguagemodelhacking}.
Here, we investigate this phenomenon with $\tps$, measuring the extent to which the inclusion of a technical definition in the context persuades the LLM toward ground-truth expert annotations.

\paragraph{Setup.} \citet{le_mens_positioning_2025} use an LLM to position sentences from the manifestos of British political parties on a five-point \answertext{left}-to-\answertext{right} scale.\footnote{Data originally annotated by \citealt{BENOIT_CONWAY_LAUDERDALE_LAVER_MIKHAYLOV_2016}}
Sentences can either relate to \emph{social} or \emph{economic} issues, and the instructions are adapted accordingly.
We filter their data to include sentences annotated by at least three experts, ensuring high agreement by limiting the standard deviation of labels to 0.5.
This process leads to 3,623 economic and 785 social sentences.
Models are instructed to code the texts from \answertext{\tone} to \answertext{\tfive}, where \answertext{\tone} is ``Extremely left'' and \answertext{\tfive} is ``Extremely right.''
Drawing from \citet{kung-peng-2023-models}, the \emph{prior} context does not include any further definition.
We use the ordinal $\dtps$ to measure the persuasiveness of two contexts: (1) one with an extensive technical definition from the original paper and (2) one with random five-shot exemplars of expert annotations (prompts in \cref{app:social_science_prompts}).
The target is the averaged expert annotation rounded to the nearest integer.
We use Qwen-2.5-7B-Instruct for all experiments.

\begin{figure}
\includegraphics[width=\columnwidth]{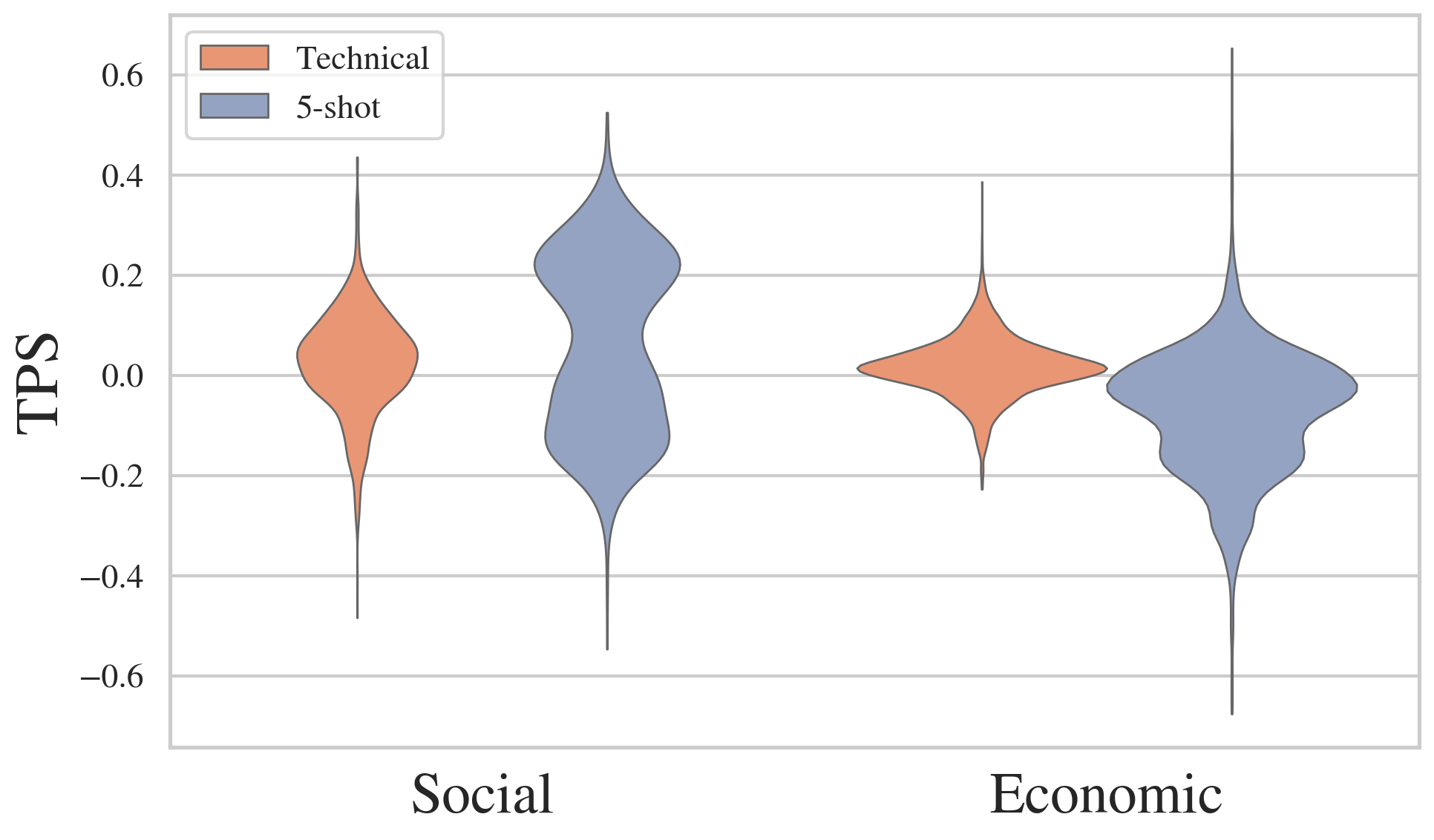}
\caption{Highly detailed prompts do not consistently influence models toward expert ratings. Here, an LM places sentences with political content on a left-right scale; the $\dtps$ measures the persuasiveness of a \textcolor[HTML]{E99675}{technical} prompt containing precise term definitions and a \textcolor[HTML]{95A3C3}{five-shot} prompt with labeled exemplars, relative to a basic prompt with little information. Sentences relate to either social ($n=785$) or economic ($n=3,623$) issues.}
\label{fig:social_exp_violin}
\end{figure}

\paragraph{Results.} While the technical definitions can alter individual responses, the net result is relatively minor: there is a small shift toward the expert definitions for both social ($\mu=0.011, \sigma^2=0.095$) and economic sentences ($\mu=0.016, \sigma^2=0.058$); see \cref{fig:social_exp_violin}.\footnote{All means are statistically different than 0 in a two-sided $t$-test; $p$-values are all near zero. An equivalent figure for $\basictps$ is in the appendix, \cref{fig:social_exp_violin_basic}.}
Importantly, although a prompt containing detailed term definitions may improve net model agreement with experts---the RMSE for the social sentences reduces from 0.99 to 0.88---$\dtps$ reveals a more nuanced picture, with inconsistent influence across examples.
The five-shot prompts induce a greater variation in responses, particularly for social sentences ($\mu=0.062, \sigma^2=0.175$); interestingly, they can also persuade the model somewhat \emph{away} from expert opinion in the case of the economic topic ($\mu=-0.072, \sigma^2=0.124$). While greater influence might be possible under a tuned set of exemplars, using a single random sample is not uncommon \cite[e.g.,][]{licht2025measuringscalarconstructssocial}.\looseness=-1

\section{Conclusion}
Our proposed metric, targeted persuasion score, measures the change in a model's probability distribution toward a target.
Promising future directions if work involve using $\tps$ to better understand how language models integrate context and prior knowledge in settings like retrieval-augmented generation and in-context learning, where we can carefully quantify the persuasiveness of specific documents or few-shot examples, the number of documents (or examples), and different orderings.

\section*{Limitations}
We face some technical limitations in executing the empirical aspects of this work.
While \Cref{sec:formalization} defines the output answer space as the set of all possible outputs $\Sigma^*$, in practice, it is computationally expensive to estimate that probability distribution. 
Instead, we look only at the model's probability distribution of the next token, which could be a noisy signal, especially in cases where the answer suggested by a context and the answer suggested from prior knowledge share the same first token.

Another limitation of $\tps$ is that this metric requires a finite answer space. This assumption is natural in settings such as multiple-choice QA, ordinal ratings, or word-sense disambiguation, where the set of possible answers can be explicitly enumerated. However, in tasks with open-ended outputs (e.g., free-form text generation), the answer space is infinite, and $\tps$ cannot be directly applied. One possible workaround is to approximate the answer space by clustering responses into a finite set of semantic categories, though we hope that better approaches for this limitation can be investigated in future work.

\section*{Contribution Statement}
Tu led the execution of the project including all coding, data processing, and analyses, building on an existing codebase from Kevin. 
The team together brainstormed and formalized the $\tps$ in terms of the Wasserstein distance, with Ryan driving the use of Wasserstein distance. 
Kevin conceptualized the initial project idea as a followup to his prior work, coordinated the project overall, proposed the distance based $\tps$, designed the lost in the middle experiments (\Cref{sec:otps_experiments}), and led writing. 
Tu and Kevin jointly designed the remaining experiments in \Cref{sec:otps_experiments}.
Kevin and Alexander jointly conceptualized the project scope and contribution. 
Alexander conceptualized the semantic distance based $\tps$. Tu formalized the semantic distance based $\tps$ and further designed the case study showing a usage of the semantic distance based $\tps$ in action. 
Alexander proposed the application study in \Cref{sec:otps_experiments}, which Tu executed. Kevin, Alexander, and Ryan advised on methodological design and conceptualization throughout. 
Ryan provided feedback and suggestions on later versions of the experimental design and carefully refined the camera-ready version of the paper.
All contributed to writing the final manuscript.

\section*{Ethics Statement}
We foresee no particular ethical concerns with this work, but hope our article contributes to developing tools that can identify and mitigate ethical concerns with LMs in the future.

\bibliography{references/custom,references/2-7-24, references/12-12-23-context-bias}

\appendix
\newpage
\onecolumn
\section{Comparison between $\tps$ and KL-divergence metric}
To further elucidate the distinction between $\tps$ and the persuasion score, based on the KL divergence, introduced by \citep{du-etal-2024-context}, consider a multiple-choice question with four options, \answertext{A},\answertext{B},\answertext{C},\answertext{D}. For example, we have the query \querytext{What is the official language of Austria}, and 4 possible answers: \answertext{A: German}, \answertext{B: English}, \answertext{C: French}, \answertext{D: Tagalog}. Suppose that initially the language model assigns a uniform distribution across the four answers.
Further suppose we have two different contexts, both intended to persuade the model toward option \answertext{A}. In the first case, the model's context-conditional distribution assigns probability 1 to \answertext{A}. In the second case, the context-conditional distribution instead assigns probability 1 to \answertext{B}. Under our $\basictps$, the first scenario yields a high positive score (0.75) because the model has been persuaded entirely toward the target answer, while the second yields a negative score (-0.25) because the model was pushed away from the target. However, the \citeposs{du-etal-2024-context}persuasion score assigns the same value to both scenarios, since in both cases the context-conditional distribution is equally different from the uniform prior distribution, and thus can not capture the effect of the first context in persuading the model toward the target answer \answertext{A}.

\section{Results on Additional Model Families}
\label{app:moremodels}

To test whether the results presented in \cref{sec:otps_experiments} generalize beyond Qwen models, we repeated the 3 $\dtps$ experiments in  \cref{sec:exp_tps_vs_k}, \cref{sec:concat_vs_individual}, and \cref{sec:lost_in_the_middle} using Llama 3B model \cite{touvron2023llamaopenefficientfoundation} and Gemma 7B models \cite{gemma_2024}. 

\begin{figure*}
    \begin{subfigure}[t]{0.32\textwidth}
        \includegraphics[width=\textwidth]{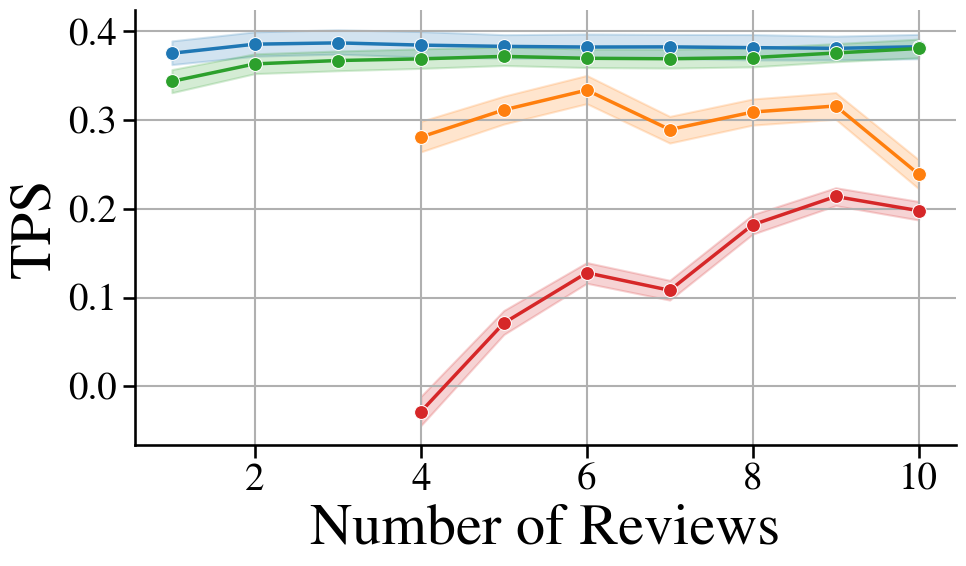}
        \caption{Qwen}
        \label{fig:otps_vs_k_qwen}
    \end{subfigure}
    \hfill
    \begin{subfigure}[t]{0.32\textwidth}
        \includegraphics[width=\textwidth]{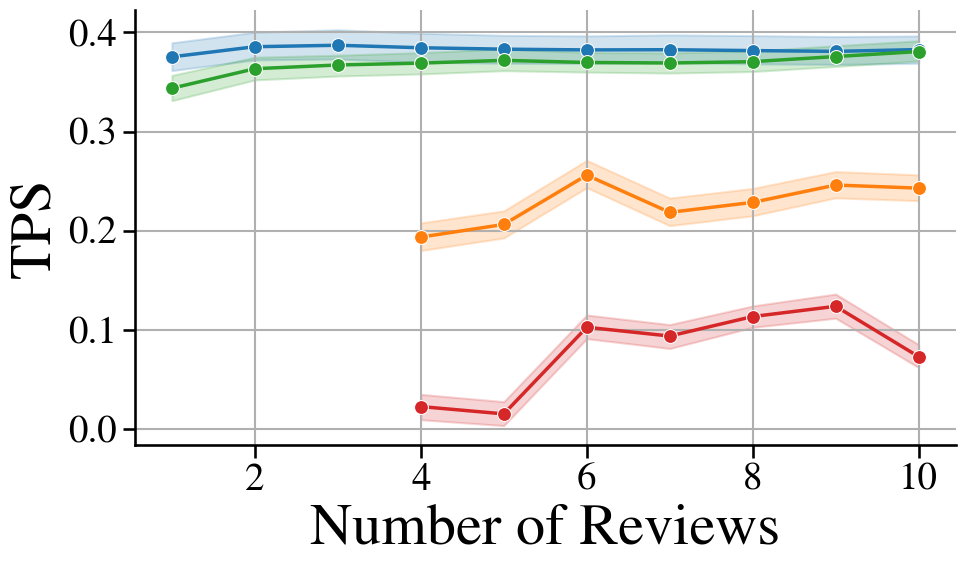}
        \caption{Llama}
        \label{fig:otps_vs_k_llama}
    \end{subfigure}
    \hfill
    \begin{subfigure}[t]{0.32\textwidth}
    \includegraphics[width=\textwidth]{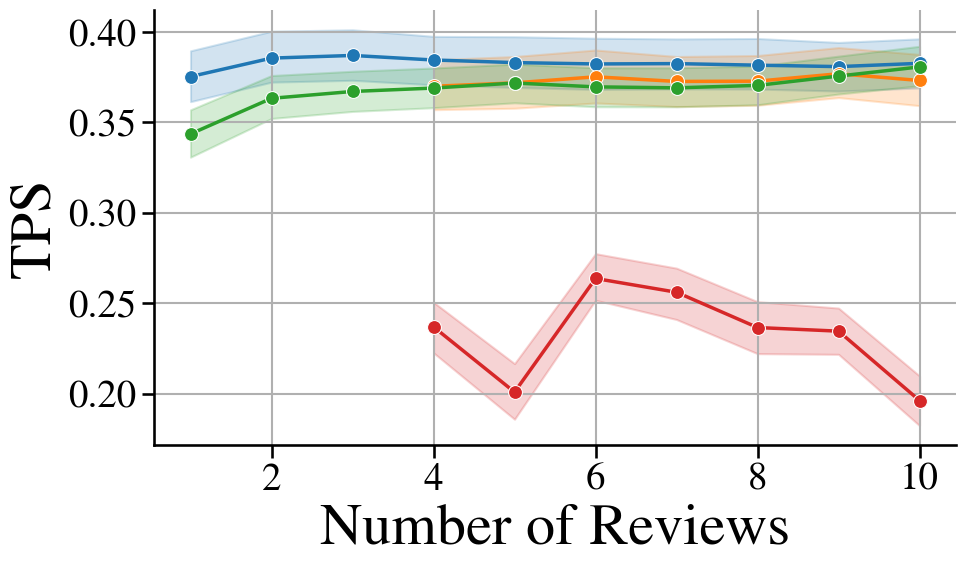}
        \caption{Gemma}
        \label{fig:otps_vs_k_gemma}
    \end{subfigure}
    \caption{How $\dtps$ varies with number of reviews distinguished by four types of reviews: \textcolor[HTML]{1f77b4}{negative}, \textcolor[HTML]{2ca02c}{positive}, \textcolor[HTML]{ff7f0e}{noisy negative}, and \textcolor[HTML]{d62728}{noisy positive}. On the left is the result of Qwen model shown in \cref{sec:exp_tps_vs_k}, followed by Llama and Gemma models' result in the middle and on the right.}
    \label{fig:otps_vs_k_generalize}
\end{figure*}

\paragraph{Effect of noise and number of context in $\dtps$} 
As in \cref{sec:exp_tps_vs_k}, we examined how $\dtps$ changes as the number of uniformly positive or negative reviews increases, as well as in noisy reviews. The results (\cref{fig:otps_vs_k_generalize}) show the same pattern as those of Qwen model: When $k$ is small, the uniformly \emph{negative} reviews have higher $\dtps$ scores than uniformly \emph{positive} reviews. Also, the effect of negative contexts having higher $\dtps$ than positive contexts disappears as the number of reviews increases. For noisy reviews, we again see that when the context is mostly positive but features some negative reviews, the $\dtps$ is much lower than when the context is mostly negative but features some positive reviews, showcasing that negative information in-context may be more persuasive than positive information in context not just in Qwen model, but also for Llama and Gemma models.

\paragraph{Concatenated vs individual reviews} 
Replicating the setup from \cref{sec:concat_vs_individual},we compared $\dtps$ of concatenated contexts to the mean of their individual sub-reviews. The qualitative results mirror those reported for Qwen in \Cref{sec:concat_vs_individual}. 
For both LLaMA and Gemma, the $\dtps$ of concatenated reviews closely matches the mean $\dtps$ of their individual reviews in the purely positive and purely negative settings, as expected. 
With noisy contexts, we again see that the noisy contexts still show a strong, monotonic relationship between the
$\dtps$ of concatenated reviews and the
mean $\dtps$ of its individual reviews, and the slope is greater than 1 again suggests that the
contradictory reviews in the concatenated contexts
are weighted less than the majority reviews, which is the same as in Qwen model.

\begin{figure*}
    \begin{subfigure}[t]{0.32\textwidth}
        \includegraphics[width=\textwidth]{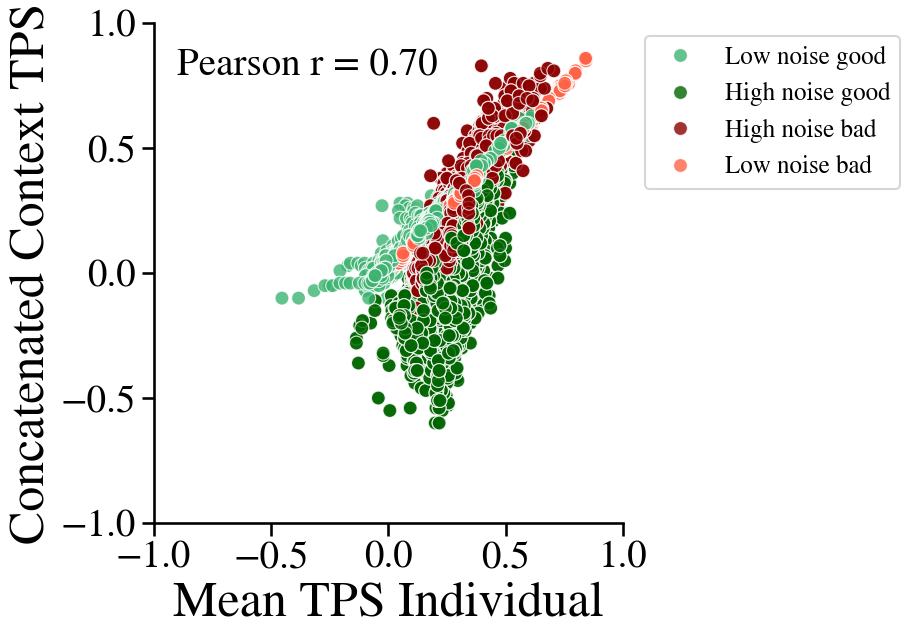}
        \caption{Qwen}
        \label{fig:tpsconcatvsmean_Qwen}
    \end{subfigure}
    \hfill
    \begin{subfigure}[t]{0.32\textwidth}
        \includegraphics[width=\textwidth]{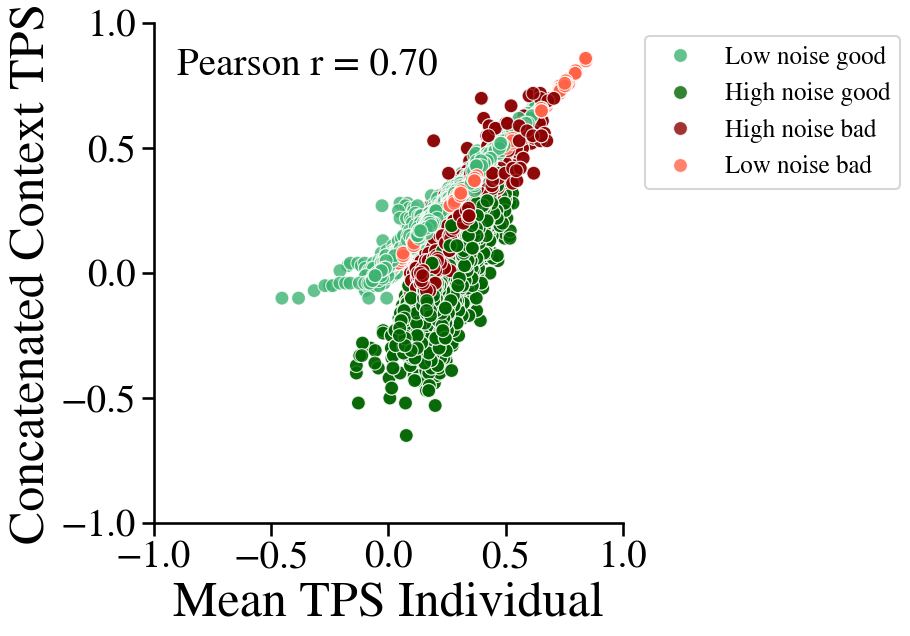}
        \caption{Llama}
        \label{fig:tpsconcatvsmean_Llama}
    \end{subfigure}
    \hfill
    \begin{subfigure}[t]{0.32\textwidth}
    \includegraphics[width=\textwidth]{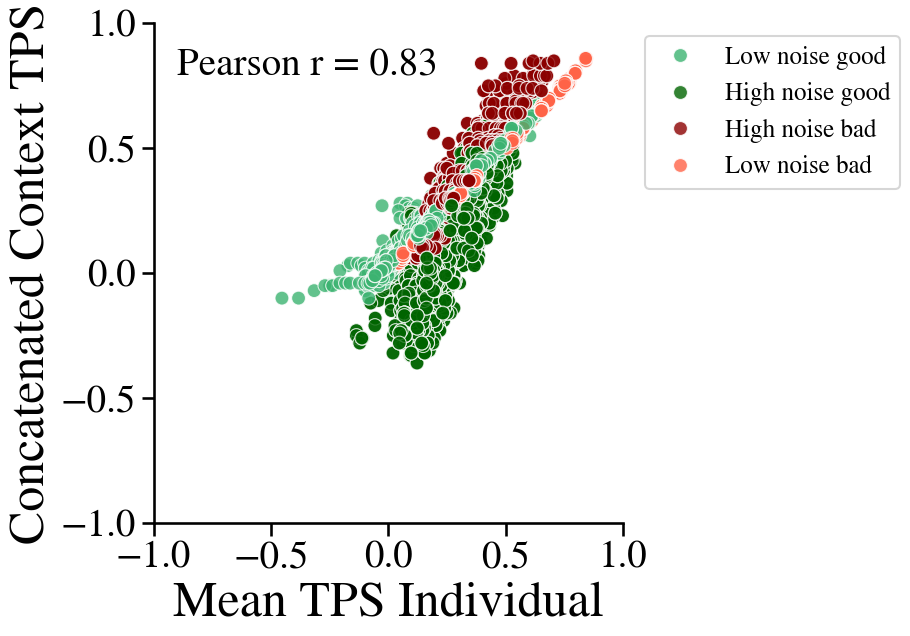}
        \caption{Gemma}
        \label{fig:tpsconcatvsmean_Gemma}
    \end{subfigure}
    \caption{$\dtps$ of concatenated review against mean $\dtps$ of its individual reviews. On the left is result with Qwen model, as in \cref{sec:concat_vs_individual}, in the middle (Llama) and on the right (Gemma), the behavior of $\dtps$ are similar to that of Qwen.}
    \label{fig:tpsconcatvsmean_generalize}
\end{figure*}

\paragraph{Lost-in-the-middle effect.} 
Following \cref{sec:lost_in_the_middle}, we tested the placement of a single contradictory review within a longer positive context. For Llama model, a significant percentage of the dataset consist of anomalies captured by the $\dtps$ but not by the decoded ratings increase. The effect is less clear in Gemma model. However, in both model family, except for Gemma-7B, through $\dtps$, one can see a “lost-in-the-middle” effect where contradictory contexts inserted at the start and end of a context influence $\dtps$ more, while such a pattern is invisible when looking at decoding-only measures.

\begin{figure*}[ht]
    \begin{subfigure}[t]{0.32\textwidth}
        \includegraphics[width=\textwidth]{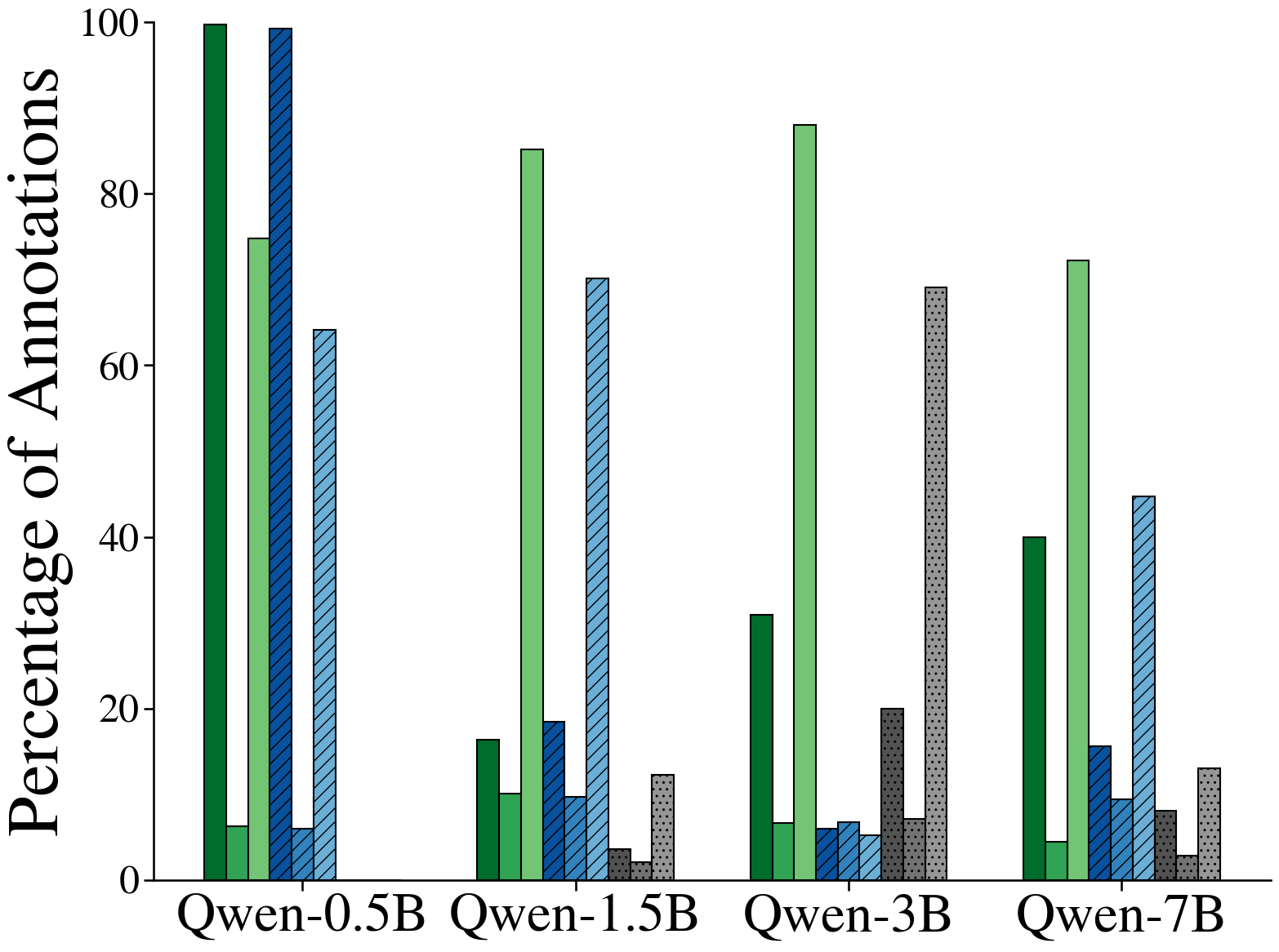}
        \caption{Qwen}
        \label{fig:otps_vs_k_qwen}
    \end{subfigure}
    \hfill
    \begin{subfigure}[t]{0.32\textwidth}
        \includegraphics[width=\textwidth]{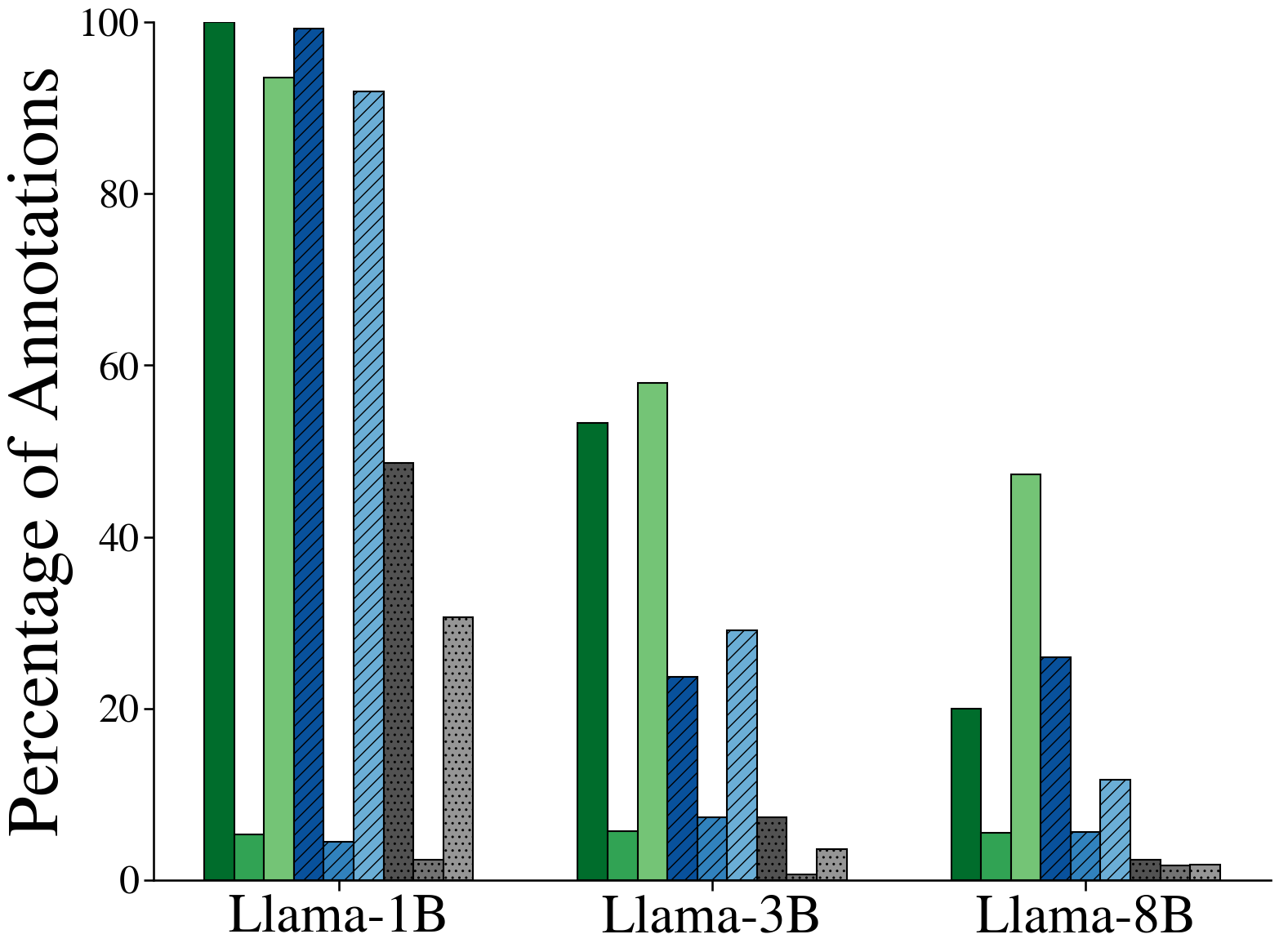}
        \caption{Llama}
        \label{fig:otps_vs_k_llama}
    \end{subfigure}
    \hfill
    \begin{subfigure}[t]{0.32\textwidth}
    \includegraphics[width=\textwidth]{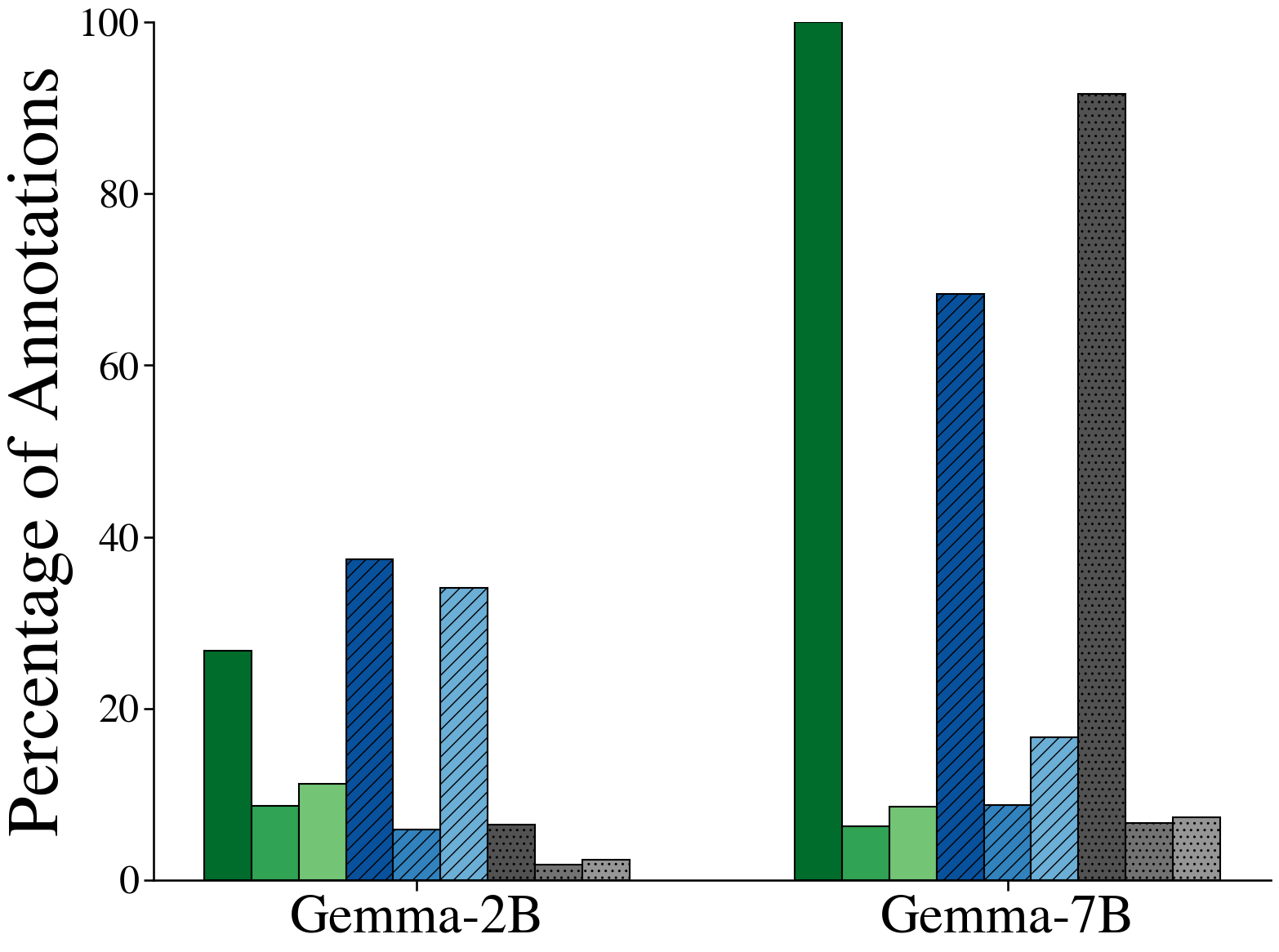}
        \caption{Gemma}
        \label{fig:otps_vs_k_gemma}
    \end{subfigure}
    \caption{}
    \label{fig:anomaly_generalize}
\end{figure*}

\section{Additional Political Positioning Result}
\begin{figure}[t]
\includegraphics[width=0.45\columnwidth]{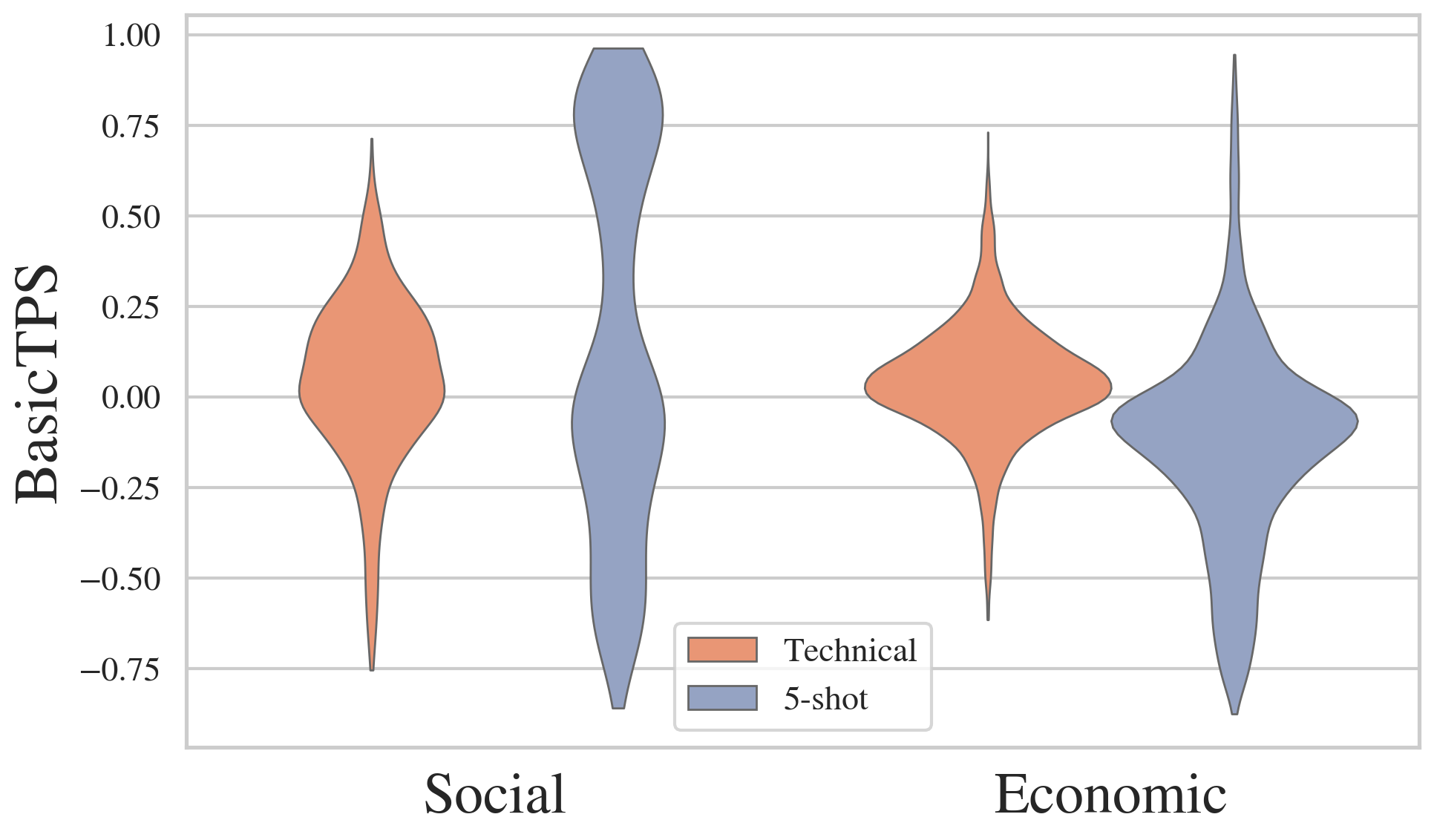}
\caption{An LLM places sentences with political content on a left-right scale. In this case, $\basictps$ measures the persuasiveness of a \textcolor[HTML]{E99675}{technical} prompt containing precise term definitions and a \textcolor[HTML]{95A3C3}{five-shot} prompt with labeled exemplars, relative to a simple prompt with little information. Sentences relate to either social ($n=785$) or economic ($n=3,623$) issues.}
\label{fig:social_exp_violin_basic}
\end{figure}

We show a version of \cref{fig:social_exp_violin} but for $\basictps$ in \cref{fig:social_exp_violin_basic}.

\section{Prompts}

\subsection{Prompts for Official Language}\label{app:official_language_prompts}
\begin{prompt}[title={Official Language Prompt.}, label=prompt:official_language]
\begin{lstlisting}
Q: What is the official language of{entity}?
A:
\end{lstlisting}
\end{prompt}

\subsection{Prompts for Movie Review}\label{app:movie_review}
\begin{prompt}[title={Movie Review Prompt Without Context}, label=prompt:movie_review_prior]
\begin{lstlisting}
Q: On a scale of 1 to 9, what is the rating of {entity}?
A:
\end{lstlisting}
\end{prompt}

\begin{prompt}[title={Example Of Movie Review Prompt With Context}, label=prompt:movie_review_posterior]
\begin{lstlisting}
Q: Here are some reviews of the film. Review 1: This film was well done. The script was tight, the direction was thoughtful. Review 2: Absolutely fantastic. I was locked in from the first scene. Review 3:  really wanted to enjoy this movie, but it lacked the spark to make it memorable. On a scale of 1 to 9, what is the rating of {entity}?
A:
\end{lstlisting}
\end{prompt}

\subsection{Prompts For Word Sense Disambiguation}
\begin{prompt}[title={Word Sense Prompt Without Context}, label=prompt:word_sense]
\begin{lstlisting}
Q: Choose among 4 definitions X, Y, Z, T 
Definition X: move fast by using one's feet 
Definition Y: direct or control a business or activity 
Definition Z: a score in baseball made by a
runner reaching home base
Definition T: a short trip or errand 
The most suitable definition for the word {entity} is (Only answer X or Y or Z or T, no more)? 
A: 
\end{lstlisting}
\end{prompt}

\begin{prompt}[title={Word Sense Prompt With Context}, label=prompt:word_sense]
\begin{lstlisting}
Context: {context sentence}
Q: Choose among 4 definitions X, Y, Z, T based on the given sentence. 
Definition X: move fast by using one's feet 
Definition Y: direct or control a business or activity 
Definition Z: a score in baseball made by a
runner reaching home base 
Definition T: a short trip or errand 
The most suitable definition for the word {entity} is (Only answer X or Y or Z or T, no more)? 
A: 
\end{lstlisting}
\end{prompt}

\subsection{Prompts for Congressional Tweet Coding}\label{app:social_science_prompts}
\begin{prompt}[title={Basic Economic Prompt.}, label=prompt:economic_basic]
\begin{lstlisting}
You will be provided with a sentence from a party manifesto. Where does this sentence stand on the "left" to "right" wing scale, in terms of economic policy? Provide your response as a score between 1 and 5 where 1 means "Extremely left" and 5 means "Extremely right". Do not provide explanations.

###
SENTENCE: {text}
LABEL: 
\end{lstlisting}
\end{prompt}

\begin{prompt}[title={Technical Economic Prompt.}, label=prompt:economic_technical]
\begin{lstlisting}
You will be provided with a sentence from a party manifesto. Where does this sentence stand on the "left" to "right" wing scale, in terms of economic policy?

**Economic** policies deal with all aspects of the economy, including:
Taxation;
Government spending;
Services provided by the government or other public bodies;
Pensions, unemployment and welfare benefits, and other state benefits;
Property, investment and share ownership, public or private;
Interest rates and exchange rates;
Regulation of economic activity, public or private;
Relations between employers, workers and trade unions.

**Left** economic policies tend to favor one or more of the following:
High levels of services provided by the government and state benefits, even if this implies high levels of taxation;
Public investment. Public ownership or control of sections of business and industry;
Public regulation of private business and economic activity;
Support for workers/trade unions relative to employers.

**Right** economic policies tend to favor one or more of the following:
Low levels of taxation, even if this implies low levels of levels of services provided by the government and state benefits;
Private investment. Minimal public ownership or control of business and industry;
Minimal public regulation of private business and economic activity;
Support for employers relative to trade unions/workers

Provide your response as a score between 1 and 5 where 1 means "Extremely left" and 5 means "Extremely right". Do not provide explanations.

###
SENTENCE: {text}
LABEL: 
\end{lstlisting}
\end{prompt}

\begin{prompt}[title={Few-shot Economic Prompt.}, label=prompt:economic_fewshot]
\begin{lstlisting}
You will be provided with a sentence from a party manifesto. Where does this sentence stand on the "left" to "right" wing scale, in terms of economic policy? Provide your response as a score between 1 and 5 where 1 means "Extremely left" and 5 means "Extremely right". Do not provide explanations.

###
SENTENCE: To achieve that, we have encouraged them to work in partnership with central government, with private enterprise, and other organisations in their community.
LABEL: 4
---
SENTENCE: The Widdicombe Report into the conduct of local authority business painted a disturbing picture of the breakdown of democratic processes in a number of councils.
LABEL: 4
---
SENTENCE: The basic rate of tax will remain unchanged at 25 per cent, as will the 40 per cent rate.
LABEL: 3
---
SENTENCE: The purpose of the review is to improve the overall prospects of students so that more are encouraged to enter higher education.
LABEL: 2
---
SENTENCE: The problem: The NHS has been squeezed between rising demand and government underfunding, and disrupted by repeated changes in government policy.
LABEL: 2
---
SENTENCE: {text}
LABEL: 
\end{lstlisting}
\end{prompt}

\begin{prompt}[title={Basic Social Prompt.}, label=prompt:social_basic]
\begin{lstlisting}
You will be provided with a sentence from a party manifesto. Where does this sentence stand on the "left" to "right" wing scale, in terms of social policy? Provide your response as a score between 1 and 5 where 1 means "Extremely left" and 5 means "Extremely right". Do not provide explanations.

###
SENTENCE: {text}
LABEL: 
\end{lstlisting}
\end{prompt}

\begin{prompt}[title={Technical Social Prompt.}, label=prompt:social_technical]
\begin{lstlisting}
You will be provided with a sentence from a party manifesto. Where does this sentence stand on the "left" to "right" wing scale, in terms of social policy?

**Social** policies deal with aspects of social and moral life, relationships between social groups, and matters of national and social identity, including:
Policing, crime, punishment and rehabilitation of offenders;
Immigration, relations between social groups, discrimination and multiculturalism;
The role of the state in regulating the social and moral behavior of individuals.
**Liberal** social policies tend to favor one or more of the following:
Recording the contents of political text on economic and social scales;
Policies emphasizing prevention of crime, rehabilitation of convicted criminals;
The right of individuals to make personal moral choices on matters such as abortion, gay rights, and euthanasia;
Policies penalizing discrimination against particular social groups and/or favoring a multicultural
society.
**Conservative** social policies tend to favor one or more of the following:
Policies emphasizing more aggressive policing, increasing police numbers, conviction and punishment of criminals, building more prisons;
The right of society to regulate personal moral choices on matters such as abortion, gay rights, and euthanasia;
Policies favoring restriction of immigration, and/or opposing explicit provision of state services for minority cultures.

Provide your response as a score between 1 and 5 where 1 means "Extremely left" and 5 means "Extremely right". Do not provide explanations.

###
SENTENCE: {text}
LABEL: 
\end{lstlisting}
\end{prompt}

\begin{prompt}[title={Few-shot Social Prompt.}, label=prompt:social_fewshot]
\begin{lstlisting}
You will be provided with a sentence from a party manifesto. Where does this sentence stand on the "left" to "right" wing scale, in terms of social policy? Provide your response as a score between 1 and 5 where 1 means "Extremely left" and 5 means "Extremely right". Do not provide explanations.

###
SENTENCE: We will also help those granted refugee status to integrate into the local community, supporting them so they can come off benefits and into work.
LABEL: 2
---
SENTENCE: We will ensure that immigration policy is non-discriminatory in its application.
LABEL: 2
---
SENTENCE: Make pensions fairer to women, by working to replace the contributory system with pension rights based on citizenship and residence in the UK.
LABEL: 2
---
SENTENCE: Britain experiences less violent crime than many comparable countries.
LABEL: 3
---
SENTENCE: As resources allow, we will increase the number of hours prisoners spend in education and training.
LABEL: 2
---
SENTENCE: {text}
LABEL: 
\end{lstlisting}
\end{prompt}

\section{Extended word sense results for \cref{sec:word-sence-case-study}}\label{app:word-sense}

\small
\setlength{\LTpre}{0pt}
\setlength{\LTpost}{0pt}

\newlength{\tpsTableW}
\setlength{\tpsTableW}{\dimexpr\linewidth-6\tabcolsep\relax}

\begin{longtable}{@{}L{0.16\tpsTableW} L{0.48\tpsTableW} C{0.18\tpsTableW} C{0.18\tpsTableW}@{}}
\toprule
\textbf{Word} & \textbf{Sense} & \textbf{Mean BasicTPS (SD)} & \textbf{Mean Distance-basedTPS (SD)} \\
\midrule
\endfirsthead

\toprule
\textbf{Word} & \textbf{Sense} & \textbf{Mean BasicTPS (SD)} & \textbf{Mean Distance-basedTPS (SD)} \\
\midrule
\endhead

\bottomrule
\endfoot
\textbf{agreement} & compatibility of observations & 0.968 (0.039) & 0.541 (0.018) \\
 & the thing arranged or agreed to & 0.186 (0.004) & 0.110 (0.002) \\
 & the statement (oral or written) of an exchange of promises & 0.927 (0.054) & 0.908 (0.045) \\
 & harmony of people's opinions or actions or characters & 0.805 (0.052) & 0.458 (0.029) \\
\midrule
\textbf{altogether} & to a complete degree or to the full or entire extent (`whole' is often used informally for `wholly') & 0.987 (0.000) & 0.967 (0.000) \\
 & with everything included or counted & 0.996 (0.000) & 0.784 (0.000) \\
 & with everything considered (and neglecting details) & 0.916 (0.001) & 0.797 (0.001) \\
 & informal terms for nakedness & 0.095 (0.030) & 0.085 (0.024) \\
\midrule
\textbf{apprize} & increase the value of & 0.971 (0.031) & 0.547 (0.024) \\
 & gain in value & 0.652 (0.008) & 0.555 (0.006) \\
 & make aware of & 0.313 (0.247) & 0.319 (0.178) \\
 & inform (somebody) of something & 0.556 (0.015) & 0.430 (0.011) \\
\midrule
\textbf{backbone} & fortitude and determination & 0.939 (0.115) & 0.486 (0.059) \\
 & the part of a book's cover that encloses the inner side of the book's pages and that faces outward when the book is shelved & 0.965 (0.039) & 0.881 (0.036) \\
 & the series of vertebrae forming the axis of the skeleton and protecting the spinal cord & 0.943 (0.078) & 0.936 (0.078) \\
 & a central cohesive source of support and stability & 0.008 (0.001) & 0.007 (0.001) \\
\midrule
\textbf{ball} & one of the two male reproductive glands that produce spermatozoa and secrete androgens & 0.996 (0.004) & 0.954 (0.004) \\
 & a solid projectile that is shot by a musket & 0.830 (0.057) & 0.628 (0.043) \\
 & a compact mass & 0.657 (0.323) & 0.107 (0.003) \\
 & an object with a spherical shape & 0.217 (0.002) & 0.100 (0.001) \\
\midrule
\textbf{beginning} & the time at which something is supposed to begin & 0.631 (0.009) & 0.014 (0.008) \\
 & the act of starting something & 0.994 (0.009) & 0.042 (0.001) \\
 & serving to begin & 0.972 (0.052) & 0.947 (0.050) \\
 & the place where something begins, where it springs into being & 0.348 (0.032) & 0.007 (0.004) \\
\midrule
\textbf{best} & the person who is most outstanding or excellent; someone who tops all others & 0.116 (0.003) & 0.108 (0.003) \\
 & get the better of & 0.880 (0.012) & 0.823 (0.011) \\
 & (comparative and superlative of `well') wiser or more advantageous and hence advisable & 0.998 (0.002) & 0.500 (0.001) \\
 & from a position of superiority or authority & 1.000 (0.001) & 0.491 (0.001) \\
\midrule
\textbf{blind} & something intended to misrepresent the true nature of an activity & 0.889 (0.068) & 0.886 (0.067) \\
 & a protective covering that keeps things out or hinders sight & 0.933 (0.202) & 0.255 (0.045) \\
 & unable to see; --Kenneth Jernigan & 0.102 (0.001) & 0.097 (0.001) \\
 & not based on reason or evidence & 0.987 (0.036) & 0.249 (0.026) \\
\midrule
\textbf{body} & the property of holding together and retaining its shape & 0.926 (0.232) & 0.937 (0.169) \\
 & the entire structure of an organism (an animal, plant, or human being) & 0.998 (0.000) & 0.729 (0.000) \\
 & a natural object consisting of a dead animal or person & 0.956 (0.059) & 0.716 (0.003) \\
 & the body excluding the head and neck and limbs & 0.030 (0.004) & 0.022 (0.003) \\
\midrule
\textbf{catalogue} & a book or pamphlet containing an enumeration of things & 0.984 (0.054) & 0.984 (0.054) \\
 & a complete list of things; usually arranged systematically & 0.998 (0.004) & 0.964 (0.004) \\
 & make a catalogue, compile a catalogue & -0.001 (0.003) & -0.001 (0.002) \\
 & make an itemized list or catalog of; classify & 0.781 (0.319) & 0.005 (0.004) \\
\midrule
\textbf{charm} & attract; cause to be enamored & -0.001 (0.001) & -0.001 (0.001) \\
 & induce into action by using one's charm & 0.983 (0.026) & 0.300 (0.009) \\
 & a verbal formula believed to have magical force & 0.993 (0.009) & 0.880 (0.008) \\
 & attractiveness that interests or pleases or stimulates & 0.978 (0.048) & 0.978 (0.048) \\
\midrule
\textbf{club} & a team of professional baseball players who play and travel together & 0.877 (0.123) & 0.770 (0.077) \\
 & a building that is occupied by a social club & 0.016 (0.011) & 0.009 (0.006) \\
 & a formal association of people with similar interests & 0.939 (0.170) & 0.524 (0.095) \\
 & a spot that is open late at night and that provides entertainment (as singers or dancers) as well as dancing and food and drink & 0.988 (0.020) & 0.879 (0.015) \\
\midrule
\textbf{compass} & an area in which something acts or operates or has power or control: & 0.953 (0.013) & 0.952 (0.013) \\
 & the limit of capability & 0.995 (0.019) & 0.673 (0.013) \\
 & get the meaning of something & 0.969 (0.067) & 0.947 (0.065) \\
 & travel around, either by plane or ship & 0.049 (0.000) & 0.048 (0.000) \\
\midrule
\textbf{consecutive} & in regular succession without gaps & 0.888 (0.141) & 0.171 (0.058) \\
 & in a consecutive manner & -0.003 (0.009) & -0.003 (0.008) \\
 & one after the other & 0.643 (0.310) & 0.541 (0.257) \\
 & successive (without a break) & 0.986 (0.034) & 0.986 (0.034) \\
\midrule
\textbf{constitute} & to compose or represent: & 0.067 (0.000) & 0.046 (0.000) \\
 & form or compose & 0.870 (0.156) & 0.759 (0.135) \\
 & create and charge with a task or function & 1.000 (0.001) & 0.458 (0.000) \\
 & set up or lay the groundwork for & 0.966 (0.001) & 0.503 (0.001) \\
\midrule
\textbf{contend} & come to terms with & 0.932 (0.090) & 0.284 (0.036) \\
 & maintain or assert & 0.965 (0.049) & 0.797 (0.046) \\
 & to make the subject of dispute, contention, or litigation & 0.039 (0.009) & 0.012 (0.003) \\
 & be engaged in a fight; carry on a fight & 0.903 (0.170) & 0.685 (0.136) \\
\midrule
\textbf{dampen} & smother or suppress & 0.590 (0.185) & 0.520 (0.164) \\
 & make moist & 0.943 (0.217) & 0.703 (0.137) \\
 & make vague or obscure or make (an image) less visible & 0.932 (0.196) & 0.284 (0.005) \\
 & lessen in force or effect & 0.323 (0.002) & 0.282 (0.002) \\
\midrule
\textbf{dependable} & worthy of reliance or trust & 0.097 (0.137) & 0.096 (0.136) \\
 & worthy of being depended on & 0.000 (0.002) & 0.001 (0.002) \\
 & financially sound & 0.972 (0.035) & 0.011 (0.013) \\
 & consistent in performance or behavior & 0.864 (0.231) & -0.017 (0.106) \\
\midrule
\textbf{die} & stop operating or functioning & 0.995 (0.010) & 0.995 (0.010) \\
 & pass from physical life and lose all bodily attributes and functions necessary to sustain life & 0.997 (0.005) & 0.027 (0.000) \\
 & lose sparkle or bouquet & 0.932 (0.078) & 0.023 (0.002) \\
 & cut or shape with a die & 0.000 (0.002) & 0.000 (0.001) \\
\midrule
\textbf{disconnected} & lacking orderly continuity & 0.998 (0.002) & 0.763 (0.001) \\
 & (music) marked by or composed of disconnected parts or sounds; cut short crisply & 0.896 (0.002) & 0.872 (0.002) \\
 & marked by sudden changes in subject and sharp transitions & 0.316 (0.003) & 0.293 (0.003) \\
 & having been divided; having the unity destroyed; -Samuel Lubell; - E.B.White & 0.780 (0.001) & 0.716 (0.001) \\
\midrule
\textbf{doctor} & restore by replacing a part or putting together what is torn or broken & 0.928 (0.214) & 0.930 (0.213) \\
 & alter and make impure, as with the intention to deceive & 0.995 (0.013) & 0.923 (0.012) \\
 & a person who holds Ph.D. degree (or the equivalent) from an academic institution & 1.000 (0.001) & 0.012 (0.000) \\
 & a licensed medical practitioner & 0.002 (0.003) & 0.002 (0.003) \\
\midrule
\textbf{draw out} & remove, usually with some force or effort; also used in an abstract sense & 0.987 (0.000) & 0.853 (0.000) \\
 & lengthen in time; cause to be or last longer & 0.115 (0.001) & 0.105 (0.001) \\
 & remove as if by suction & 0.983 (0.058) & 0.949 (0.049) \\
 & deduce (a principle) or construe (a meaning) & 0.886 (0.054) & 0.807 (0.026) \\
\midrule
\textbf{edge} & provide with a border or edge & -0.022 (0.075) & -0.000 (0.001) \\
 & advance slowly, as if by inches & 0.995 (0.006) & 0.370 (0.002) \\
 & lie adjacent to another or share a boundary & 0.995 (0.017) & 0.008 (0.000) \\
 & the attribute of urgency in tone of voice & 0.998 (0.004) & 0.968 (0.004) \\
\midrule
\textbf{effect} & (of a law) having legal validity & 0.997 (0.005) & 0.677 (0.002) \\
 & a phenomenon that follows and is caused by some previous phenomenon & 0.346 (0.086) & 0.346 (0.086) \\
 & an outward appearance & 0.970 (0.085) & 0.803 (0.059) \\
 & produce & 0.620 (0.009) & 0.619 (0.009) \\
\midrule
\textbf{evidence} & give evidence & 0.155 (0.243) & 0.088 (0.138) \\
 & your basis for belief or disbelief; knowledge on which to base belief & 0.965 (0.126) & 0.762 (0.099) \\
 & provide evidence for & 0.007 (0.005) & 0.004 (0.003) \\
 & provide evidence for; stand as proof of; show by one's behavior, attitude, or external attributes & 0.875 (0.132) & 0.462 (0.069) \\
\midrule
\textbf{father} & the founder of a family & 0.997 (0.002) & 0.337 (0.001) \\
 & a male parent (also used as a term of address to your father) & 0.002 (0.002) & 0.000 (0.002) \\
 & a person who founds or establishes some institution & 0.809 (0.281) & 0.335 (0.162) \\
 & make children & 0.843 (0.285) & 0.852 (0.267) \\
\midrule
\textbf{fault} & a wrong action attributable to bad judgment or ignorance or inattention & 0.020 (0.003) & 0.019 (0.003) \\
 & (geology) a crack in the earth's crust resulting from the displacement of one side with respect to the other & 0.993 (0.010) & 0.986 (0.005) \\
 & the quality of being inadequate or falling short of perfection & 0.998 (0.002) & 0.792 (0.001) \\
 & an imperfection in an object or machine & 0.975 (0.010) & 0.914 (0.007) \\
\midrule
\textbf{favor} & consider as the favorite & 0.971 (0.078) & 0.435 (0.035) \\
 & an advantage to the benefit of someone or something & 0.971 (0.007) & 0.964 (0.007) \\
 & an inclination to approve & 0.998 (0.001) & 0.971 (0.001) \\
 & promote over another & 0.027 (0.004) & 0.027 (0.002) \\
\midrule
\textbf{gap} & an open or empty space in or between things & 0.168 (0.002) & 0.163 (0.002) \\
 & a conspicuous disparity or difference as between two figures & 0.821 (0.010) & 0.800 (0.010) \\
 & a narrow opening & 0.996 (0.011) & 0.804 (0.011) \\
 & an act of delaying or interrupting the continuity & 0.998 (0.002) & 0.994 (0.002) \\
\midrule
\textbf{get together} & get people together & -0.002 (0.003) & -0.002 (0.003) \\
 & a small informal social gathering & 0.885 (0.263) & 0.846 (0.256) \\
 & work together on a common enterprise of project & 0.943 (0.172) & 0.606 (0.110) \\
 & become part of; become a member of a group or organization & 0.681 (0.345) & 0.233 (0.119) \\
\midrule
\textbf{go off} & happen in a particular manner & 0.815 (0.366) & 0.780 (0.207) \\
 & burst inward & 0.684 (0.402) & 0.651 (0.279) \\
 & run away; usually includes taking something or somebody along & 0.219 (0.014) & 0.216 (0.014) \\
 & go off or discharge & 0.744 (0.047) & 0.734 (0.046) \\
\midrule
\textbf{guarantee} & make certain of & 0.834 (0.251) & 0.001 (0.007) \\
 & promise to do or accomplish & 0.706 (0.339) & 0.700 (0.336) \\
 & stand behind and guarantee the quality, accuracy, or condition of & -0.005 (0.008) & -0.002 (0.003) \\
 & give surety or assume responsibility & 0.943 (0.194) & 0.008 (0.002) \\
\midrule
\textbf{gush} & issue in a jet; come out in a jet; stream or spring forth & 0.009 (0.016) & 0.007 (0.013) \\
 & gush forth in a sudden stream or jet & -0.001 (0.001) & -0.001 (0.001) \\
 & a sudden rapid flow (as of water) & 0.002 (0.001) & 0.002 (0.001) \\
 & praise enthusiastically & 0.925 (0.227) & 0.800 (0.196) \\
\midrule
\textbf{hell} & noisy and unrestrained mischief & 0.971 (0.034) & 0.026 (0.001) \\
 & violent and excited activity & 0.989 (0.010) & 0.989 (0.010) \\
 & any place of pain and turmoil & 0.001 (0.002) & -0.000 (0.000) \\
 & a cause of difficulty and suffering & 0.717 (0.255) & 0.014 (0.005) \\
\midrule
\textbf{hide} & make undecipherable or imperceptible by obscuring or concealing & 0.992 (0.020) & 0.853 (0.017) \\
 & cover as if with a shroud & 0.996 (0.006) & 0.926 (0.006) \\
 & prevent from being seen or discovered & 0.961 (0.006) & 0.011 (0.002) \\
 & be or go into hiding; keep out of sight, as for protection and safety & 0.031 (0.001) & -0.000 (0.001) \\
\midrule
\textbf{hike} & increase & 0.795 (0.022) & 0.118 (0.020) \\
 & an increase in cost & 0.868 (0.161) & 0.807 (0.160) \\
 & the amount a salary is increased & 0.990 (0.028) & 0.813 (0.025) \\
 & a long walk usually for exercise or pleasure & 0.191 (0.010) & 0.030 (0.002) \\
\midrule
\textbf{honorable} & worthy of being honored; entitled to honor and respect & 0.406 (0.009) & 0.290 (0.007) \\
 & adhering to ethical and moral principles & 0.511 (0.204) & 0.365 (0.146) \\
 & not disposed to cheat or defraud; not deceptive or fraudulent & 0.941 (0.216) & 0.398 (0.088) \\
 & deserving of esteem and respect & 0.081 (0.092) & -0.108 (0.095) \\
\midrule
\textbf{ice} & decorate with frosting & 0.828 (0.028) & 0.768 (0.028) \\
 & water frozen in the solid state & 0.633 (0.048) & 0.455 (0.048) \\
 & a rink with a floor of ice for ice hockey or ice skating & 0.510 (0.019) & 0.361 (0.016) \\
 & diamonds & 0.948 (0.072) & 0.417 (0.063) \\
\midrule
\textbf{identical} & being the exact same one; not any other: & 0.066 (0.000) & 0.054 (0.000) \\
 & exactly alike; incapable of being perceived as different & 0.985 (0.005) & 0.086 (0.000) \\
 & (of twins) derived from a single egg or ovum & 0.943 (0.002) & 0.919 (0.002) \\
 & coinciding exactly when superimposed & 0.978 (0.089) & 0.113 (0.005) \\
\midrule
\textbf{incidental} & (frequently plural) an expense not budgeted or not specified & 0.429 (0.000) & 0.222 (0.000) \\
 & (sometimes followed by `to') minor or casual or subordinate in significance or nature or occurring as a chance concomitant or consequence & 0.554 (0.021) & 0.287 (0.011) \\
 & not of prime or central importance; - Pubs.MLA & 0.996 (0.005) & 0.382 (0.001) \\
 & following or accompanying as a consequence & 0.632 (0.344) & 0.594 (0.330) \\
\midrule
\textbf{incompetent} & not doing a good job & 0.821 (0.121) & 0.694 (0.101) \\
 & legally not qualified or sufficient & 0.987 (0.012) & 0.983 (0.012) \\
 & not meeting requirements & 0.913 (0.132) & 0.792 (0.112) \\
 & showing lack of skill or aptitude & -0.095 (0.078) & -0.080 (0.066) \\
\midrule
\textbf{indicate} & indicate a place, direction, person, or thing; either spatially or figuratively & 0.217 (0.007) & 0.002 (0.002) \\
 & suggest the necessity of an intervention; in medicine & 0.777 (0.002) & 0.007 (0.002) \\
 & be a signal for or a symptom of & 0.996 (0.004) & 0.986 (0.004) \\
 & give evidence of & 0.961 (0.077) & 0.875 (0.074) \\
\midrule
\textbf{insert} & put or introduce into something & 0.633 (0.001) & 0.466 (0.001) \\
 & fit snugly into & 0.680 (0.294) & 0.673 (0.252) \\
 & insert casually & 0.349 (0.016) & 0.256 (0.012) \\
 & introduce & 0.581 (0.364) & 0.385 (0.147) \\
\midrule
\textbf{inspire} & heighten or intensify & 0.034 (0.000) & 0.027 (0.000) \\
 & serve as the inciting cause of & 0.990 (0.005) & 0.649 (0.003) \\
 & draw in (air) & 0.919 (0.189) & 0.793 (0.162) \\
 & spur on or encourage especially by cheers and shouts & 0.965 (0.015) & 0.819 (0.013) \\
\midrule
\textbf{intend} & denote or connote & 0.336 (0.342) & -0.032 (0.153) \\
 & design or destine & 0.998 (0.006) & 0.996 (0.006) \\
 & have in mind as a purpose & 0.956 (0.194) & 0.943 (0.191) \\
 & mean or intend to express or convey & 0.002 (0.000) & 0.002 (0.000) \\
\midrule
\textbf{interrupt} & interfere in someone else's activity & 0.761 (0.268) & 0.648 (0.229) \\
 & destroy the peace or tranquility of & 0.987 (0.013) & 0.960 (0.012) \\
 & terminate & 0.475 (0.341) & 0.474 (0.341) \\
 & make a break in & 0.001 (0.003) & 0.000 (0.003) \\
\midrule
\textbf{invigorate} & impart vigor, strength, or vitality to & 0.235 (0.008) & 0.173 (0.005) \\
 & make lively & 0.990 (0.002) & 0.705 (0.002) \\
 & give life or energy to & 0.803 (0.170) & 0.776 (0.168) \\
 & heighten or intensify & 0.822 (0.001) & 0.545 (0.000) \\
\midrule
\textbf{judge} & judge tentatively or form an estimate of (quantities or time) & 0.994 (0.013) & 0.928 (0.007) \\
 & form a critical opinion of & 0.896 (0.205) & 0.795 (0.120) \\
 & pronounce judgment on & 0.065 (0.132) & 0.050 (0.106) \\
 & put on trial or hear a case and sit as the judge at the trial of & 0.824 (0.075) & 0.647 (0.059) \\
\midrule
\textbf{law} & a generalization that describes recurring facts or events in nature & 0.983 (0.063) & 0.030 (0.002) \\
 & the force of policemen and officers & 0.824 (0.225) & 0.825 (0.224) \\
 & the collection of rules imposed by authority & 0.992 (0.000) & 0.026 (0.000) \\
 & the learned profession that is mastered by graduate study in a law school and that is responsible for the judicial system & 0.005 (0.003) & 0.000 (0.000) \\
\midrule
\textbf{lecture} & deliver a lecture or talk & -0.000 (0.000) & -0.000 (0.000) \\
 & censure severely or angrily & 0.980 (0.040) & 0.896 (0.037) \\
 & a speech that is open to the public & 0.938 (0.198) & 0.305 (0.064) \\
 & a lengthy rebuke & 0.803 (0.282) & 0.184 (0.222) \\
\midrule
\textbf{lonely} & marked by dejection from being alone & 0.982 (0.038) & 0.879 (0.034) \\
 & lacking companions or companionship & 0.001 (0.002) & 0.001 (0.002) \\
 & characterized by or preferring solitude & 0.873 (0.147) & 0.873 (0.147) \\
 & devoid of creatures & 0.864 (0.226) & 0.784 (0.206) \\
\midrule
\textbf{lose} & be set at a disadvantage & 0.993 (0.021) & 0.825 (0.018) \\
 & fail to make money in a business; make a loss or fail to profit & 0.001 (0.000) & 0.001 (0.000) \\
 & fail to perceive or to catch with the senses or the mind & 0.987 (0.036) & 0.471 (0.017) \\
 & place (something) where one cannot find it again & 0.998 (0.004) & 0.614 (0.003) \\
\midrule
\textbf{modest} & limited in size or scope & 0.997 (0.005) & 0.825 (0.005) \\
 & not large but sufficient in size or amount & 0.998 (0.003) & 0.655 (0.002) \\
 & low or inferior in station or quality & 0.382 (0.001) & 0.382 (0.001) \\
 & humble in spirit or manner; suggesting retiring mildness or even cowed submissiveness & 0.603 (0.019) & 0.603 (0.019) \\
\midrule
\textbf{nice} & exhibiting courtesy and politeness & 0.955 (0.075) & 0.780 (0.062) \\
 & excessively fastidious and easily disgusted & 0.991 (0.014) & 0.804 (0.011) \\
 & done with delicacy and skill & 0.796 (0.230) & 0.498 (0.177) \\
 & socially or conventionally correct; refined or virtuous & 0.001 (0.004) & 0.001 (0.003) \\
\midrule
\textbf{nigh} & (of actions or states) slightly short of or not quite accomplished; all but & 0.991 (0.009) & 0.591 (0.005) \\
 & being on the left side & 0.999 (0.001) & 0.512 (0.001) \\
 & not far distant in time or space or degree or circumstances & 0.927 (0.166) & 0.510 (0.091) \\
 & near in time or place or relationship & -0.000 (0.001) & -0.000 (0.001) \\
\midrule
\textbf{offend} & act in disregard of laws, rules, contracts, or promises & 0.347 (0.018) & 0.291 (0.016) \\
 & strike with disgust or revulsion & 0.867 (0.124) & 0.612 (0.101) \\
 & hurt the feelings of & 0.656 (0.021) & 0.543 (0.017) \\
 & cause to feel resentment or indignation & 0.987 (0.021) & 0.902 (0.021) \\
\midrule
\textbf{orbit} & move in an orbit & -0.076 (0.267) & -0.065 (0.230) \\
 & an area in which something acts or operates or has power or control: & 0.945 (0.217) & 0.843 (0.099) \\
 & the (usually elliptical) path described by one celestial body in its revolution about another & 0.896 (0.000) & 0.770 (0.000) \\
 & a particular environment or walk of life & 0.946 (0.200) & 0.914 (0.088) \\
\midrule
\textbf{passage} & a journey usually by ship & 0.996 (0.005) & 0.662 (0.005) \\
 & a bodily reaction of changing from one place or stage to another & 0.986 (0.002) & 0.686 (0.001) \\
 & the motion of one object relative to another & 0.313 (0.003) & 0.312 (0.003) \\
 & a path or channel or duct through or along which something may pass & 0.689 (0.022) & 0.676 (0.022) \\
\midrule
\textbf{pitter-patter} & describing a rhythmic beating & 0.877 (0.198) & 0.872 (0.198) \\
 & rain gently & -0.007 (0.028) & -0.004 (0.023) \\
 & make light, rapid and repeated sounds & 0.954 (0.058) & 0.761 (0.046) \\
 & as of footsteps & 0.980 (0.032) & 0.969 (0.031) \\
\midrule
\textbf{plastic} & forming or capable of forming or molding or fashioning & 0.936 (0.185) & 0.937 (0.184) \\
 & capable of being influenced or formed & 0.957 (0.066) & 0.789 (0.044) \\
 & a card (usually plastic) that assures a seller that the person using it has a satisfactory credit rating and that the issuer will see to it that the seller receives payment for the merchandise delivered & -0.001 (0.001) & -0.001 (0.001) \\
 & capable of being molded or modeled (especially of earth or clay or other soft material) & 0.996 (0.002) & 0.754 (0.002) \\
\midrule
\textbf{pliable} & susceptible to being led or directed & -0.013 (0.204) & -0.013 (0.191) \\
 & capable of being shaped or bent or drawn out & 0.968 (0.022) & 0.853 (0.019) \\
 & able to adjust readily to different conditions & 0.999 (0.001) & 0.941 (0.001) \\
 & capable of being bent or flexed or twisted without breaking & 0.872 (0.133) & 0.809 (0.131) \\
\midrule
\textbf{pluck} & pull or pull out sharply & 0.088 (0.002) & 0.002 (0.000) \\
 & pull lightly but sharply with a plucking motion & 0.741 (0.205) & 0.021 (0.006) \\
 & strip of feathers & 0.975 (0.051) & 0.022 (0.001) \\
 & look for and gather & 0.963 (0.082) & 0.956 (0.082) \\
\midrule
\textbf{plunder} & destroy and strip of its possession & 0.691 (0.368) & 0.582 (0.332) \\
 & steal goods; take as spoils & 0.994 (0.014) & 0.864 (0.009) \\
 & plunder (a town) after capture & 0.346 (0.007) & 0.322 (0.007) \\
 & take illegally; of intellectual property & 0.644 (0.001) & 0.599 (0.001) \\
\midrule
\textbf{plunk} & set (something or oneself) down with or as if with a noise & 0.777 (0.004) & 0.560 (0.003) \\
 & drop steeply & 0.216 (0.032) & 0.158 (0.023) \\
 & with a short hollow thud & 0.987 (0.030) & 0.897 (0.030) \\
 & pull lightly but sharply with a plucking motion & 0.998 (0.002) & 0.805 (0.002) \\
\midrule
\textbf{prevail} & be valid, applicable, or true & -0.000 (0.003) & -0.000 (0.001) \\
 & continue to exist & 0.947 (0.099) & 0.003 (0.000) \\
 & be larger in number, quantity, power, status or importance & 0.999 (0.002) & 0.987 (0.002) \\
 & prove superior & 0.997 (0.006) & 0.004 (0.000) \\
\midrule
\textbf{rag} & censure severely or angrily & 0.962 (0.092) & 0.902 (0.092) \\
 & treat cruelly & 0.608 (0.029) & 0.474 (0.023) \\
 & cause annoyance in; disturb, especially by minor irritations & 0.869 (0.207) & 0.755 (0.178) \\
 & harass with persistent criticism or carping & 0.383 (0.015) & 0.300 (0.012) \\
\midrule
\textbf{rattling} & quick and energetic & 0.535 (0.408) & 0.573 (0.310) \\
 & extraordinarily good or great ; used especially as intensifiers & 0.999 (0.003) & 0.873 (0.002) \\
 & a rapid series of short loud sounds (as might be heard with a stethoscope in some types of respiratory disorders) & 0.253 (0.095) & 0.187 (0.070) \\
 & used as intensifiers; `real' is sometimes used informally for `really'; `rattling' is informal & 0.696 (0.019) & 0.515 (0.014) \\
\midrule
\textbf{refer} & be relevant to & 0.068 (0.031) & 0.025 (0.015) \\
 & make reference to & 0.769 (0.166) & 0.298 (0.066) \\
 & have as a meaning & 0.987 (0.025) & 0.681 (0.017) \\
 & seek information from & 0.924 (0.195) & 0.417 (0.096) \\
\midrule
\textbf{relegate} & refer to another person for decision or judgment & 0.847 (0.259) & 0.673 (0.206) \\
 & expel, as if by official decree & 0.901 (0.247) & 0.900 (0.247) \\
 & assign to a lower position; reduce in rank & 0.032 (0.007) & 0.026 (0.005) \\
 & assign to a class or kind & 0.551 (0.416) & 0.431 (0.331) \\
\midrule
\textbf{scourge} & a person who inspires fear or dread & 0.993 (0.020) & 0.968 (0.020) \\
 & whip & 0.068 (0.197) & 0.025 (0.194) \\
 & cause extensive destruction or ruin utterly & 0.809 (0.027) & 0.582 (0.019) \\
 & something causing misery or death & 0.997 (0.006) & 0.968 (0.006) \\
\midrule
\textbf{silly} & inspiring scornful pity; - Dashiell Hammett & 0.957 (0.103) & 0.941 (0.103) \\
 & lacking seriousness; given to frivolity & 0.720 (0.298) & 0.634 (0.262) \\
 & ludicrous, foolish & 0.092 (0.001) & 0.081 (0.001) \\
 & dazed from or as if from repeated blows & 0.525 (0.381) & 0.487 (0.357) \\
\midrule
\textbf{slant} & degree of deviation from a horizontal plane & 0.628 (0.349) & 0.476 (0.267) \\
 & to incline or bend from a vertical position & -0.000 (0.000) & -0.000 (0.000) \\
 & heel over & 0.844 (0.226) & 0.380 (0.102) \\
 & present with a bias & 0.997 (0.005) & 0.551 (0.003) \\
\midrule
\textbf{slide} & the act of moving smoothly along a surface while remaining in contact with it & 0.986 (0.058) & 0.734 (0.043) \\
 & to pass or move unobtrusively or smoothly & 0.009 (0.001) & 0.006 (0.000) \\
 & (music) rapid sliding up or down the musical scale & 0.997 (0.006) & 0.710 (0.005) \\
 & move obliquely or sideways, usually in an uncontrolled manner & 0.986 (0.004) & 0.630 (0.003) \\
\midrule
\textbf{slight} & being of delicate or slender build; - Frank Norris & 0.995 (0.007) & 0.868 (0.004) \\
 & (quantifier used with mass nouns) small in quantity or degree; not much or almost none or (with `a') at least some & 0.007 (0.001) & 0.006 (0.001) \\
 & lacking substance or significance; ; ; ; a fragile claim to fame" & 0.998 (0.001) & 0.952 (0.001) \\
 & pay no attention to, disrespect & 0.996 (0.000) & 0.995 (0.000) \\
\midrule
\textbf{smack} & directly & 0.495 (0.394) & 0.478 (0.393) \\
 & deliver a hard blow to & 0.992 (0.008) & 0.007 (0.000) \\
 & have a distinctive or characteristic taste & 0.796 (0.218) & 0.004 (0.001) \\
 & have an element suggestive (of something) & -0.005 (0.049) & 0.000 (0.000) \\
\midrule
\textbf{smother} & deprive of the oxygen necessary for combustion & 0.925 (0.070) & 0.690 (0.052) \\
 & conceal or hide & 0.829 (0.240) & 0.345 (0.146) \\
 & envelop completely & 0.988 (0.017) & 0.843 (0.015) \\
 & deprive of oxygen and prevent from breathing & 0.002 (0.001) & 0.001 (0.001) \\
\midrule
\textbf{so} & subsequently or soon afterward (often used as sentence connectors) & 0.993 (0.005) & 0.308 (0.002) \\
 & in the way indicated; ; ; (`thusly' is a nonstandard variant) & -0.184 (0.350) & -0.101 (0.195) \\
 & in truth (often tends to intensify) & 0.995 (0.002) & 0.992 (0.002) \\
 & (used to introduce a logical conclusion) from that fact or reason or as a result & 0.981 (0.004) & 0.549 (0.002) \\
\midrule
\textbf{socialize} & prepare for social life & 0.995 (0.011) & 0.684 (0.007) \\
 & make conform to socialist ideas and philosophies & 0.984 (0.050) & 0.984 (0.050) \\
 & take part in social activities; interact with others & -0.001 (0.003) & -0.001 (0.002) \\
 & train for a social environment & 0.734 (0.301) & 0.632 (0.101) \\
\midrule
\textbf{specialise} & become more focus on an area of activity or field of study & 0.908 (0.081) & 0.028 (0.002) \\
 & devote oneself to a special area of work & 0.005 (0.002) & 0.001 (0.000) \\
 & be specific about & 0.838 (0.196) & 0.814 (0.192) \\
 & suit to a special purpose & 0.994 (0.010) & 0.022 (0.000) \\
\midrule
\textbf{specialize} & become more focus on an area of activity or field of study & 0.913 (0.091) & 0.028 (0.003) \\
 & suit to a special purpose & 0.994 (0.013) & 0.021 (0.000) \\
 & be specific about & 0.797 (0.233) & 0.773 (0.230) \\
 & devote oneself to a special area of work & 0.004 (0.002) & 0.001 (0.000) \\
\midrule
\textbf{standard} & the ideal in terms of which something can be judged & 0.686 (0.223) & 0.596 (0.181) \\
 & a basis for comparison; a reference point against which other things can be evaluated & 0.251 (0.002) & 0.128 (0.001) \\
 & regularly and widely used or sold & 0.930 (0.067) & 0.399 (0.027) \\
 & conforming to the established language usage of educated native speakers;  (American);  (British) & 0.791 (0.003) & 0.389 (0.001) \\
\midrule
\textbf{stark} & devoid of any qualifications or disguise or adornment & 0.961 (0.106) & 0.668 (0.074) \\
 & severely simple & 0.001 (0.003) & 0.001 (0.003) \\
 & providing no shelter or sustenance & 0.997 (0.005) & 0.562 (0.003) \\
 & without qualification; used informally as (often pejorative) intensifiers & 0.936 (0.091) & 0.779 (0.076) \\
\midrule
\textbf{steamroller} & make level or flat with a steamroller & 0.734 (0.217) & 0.734 (0.218) \\
 & overwhelm by using great force & 0.949 (0.080) & 0.155 (0.003) \\
 & proceed with great force & 0.971 (0.034) & 0.151 (0.003) \\
 & bring to a specified state by overwhelming force or pressure & 0.132 (0.003) & 0.132 (0.002) \\
\midrule
\textbf{strangle} & kill by squeezing the throat of so as to cut off the air & -0.001 (0.001) & -0.001 (0.001) \\
 & struggle for breath; have insufficient oxygen intake & 0.393 (0.348) & 0.272 (0.240) \\
 & prevent the progress or free movement of & 0.932 (0.116) & 0.769 (0.096) \\
 & conceal or hide & 0.582 (0.297) & 0.456 (0.235) \\
\midrule
\textbf{suggest} & suggest the necessity of an intervention; in medicine & 0.984 (0.002) & 0.777 (0.002) \\
 & make a proposal, declare a plan for something & 0.965 (0.037) & 0.668 (0.026) \\
 & imply as a possibility & 0.025 (0.001) & 0.018 (0.001) \\
 & call to mind & 0.643 (0.303) & 0.024 (0.011) \\
\midrule
\textbf{sway} & win approval or support for & 0.814 (0.000) & 0.802 (0.000) \\
 & move or walk in a swinging or swaying manner & 0.179 (0.011) & 0.177 (0.011) \\
 & cause to move back and forth & 0.995 (0.008) & 0.907 (0.008) \\
 & move back and forth or sideways & 0.947 (0.210) & 0.883 (0.195) \\
\midrule
\textbf{tape} & register electronically & 0.990 (0.054) & 0.813 (0.044) \\
 & memory device consisting of a long thin plastic strip coated with iron oxide; used to record audio or video signals or to store computer information & 0.996 (0.008) & 0.599 (0.008) \\
 & measuring instrument consisting of a narrow strip (cloth or metal) marked in inches or centimeters and used for measuring lengths & 0.987 (0.035) & -0.004 (0.028) \\
 & a recording made on magnetic tape & 0.003 (0.001) & 0.002 (0.001) \\
\midrule
\textbf{terrible} & extreme in degree or extent or amount or impact & 0.082 (0.012) & 0.080 (0.012) \\
 & causing fear or dread or terror & 0.974 (0.023) & 0.718 (0.016) \\
 & exceptionally bad or displeasing & 0.989 (0.015) & 0.639 (0.010) \\
 & intensely or extremely bad or unpleasant in degree or quality & 0.864 (0.031) & 0.864 (0.029) \\
\midrule
\textbf{total} & complete in extent or degree and in every particular & 0.011 (0.001) & 0.009 (0.001) \\
 & determine the sum of & 0.348 (0.216) & 0.920 (0.088) \\
 & add up in number or quantity & 0.988 (0.022) & 0.989 (0.017) \\
 & constituting the full quantity or extent; complete & 0.949 (0.090) & 0.427 (0.064) \\
\midrule
\textbf{transplant} & lift and reset in another soil or situation & 0.018 (0.001) & 0.013 (0.000) \\
 & transfer from one place or period to another & 0.929 (0.135) & 0.636 (0.119) \\
 & an operation moving an organ from one organism (the donor) to another (the recipient) & 0.999 (0.001) & 0.145 (0.000) \\
 & the act of removing something from one location and introducing it in another location & 0.926 (0.218) & 0.143 (0.006) \\
\midrule
\textbf{unknown} & not known before & 0.976 (0.002) & 0.790 (0.002) \\
 & an unknown and unexplored region & 0.034 (0.001) & 0.026 (0.001) \\
 & not famous or acclaimed & 0.935 (0.093) & 0.927 (0.088) \\
 & being or having an unknown or unnamed source & 0.985 (0.014) & 0.419 (0.006) \\
\midrule
\textbf{vaporise} & cause to change into a vapor & 0.054 (0.001) & 0.000 (0.000) \\
 & change into a vapor & 0.263 (0.257) & -0.033 (0.058) \\
 & turn into gas & 0.953 (0.077) & 0.945 (0.076) \\
 & lose or cause to lose liquid by vaporization leaving a more concentrated residue & 0.935 (0.010) & 0.008 (0.000) \\
\midrule
\textbf{vaporize} & decrease rapidly and disappear & 0.940 (0.215) & 0.659 (0.166) \\
 & lose or cause to lose liquid by vaporization leaving a more concentrated residue & 0.994 (0.014) & 0.907 (0.014) \\
 & turn into gas & 0.417 (0.100) & 0.283 (0.068) \\
 & kill with or as if with a burst of gunfire or electric current or as if by shooting & 0.549 (0.000) & 0.373 (0.000) \\
\midrule
\textbf{verbalise} & be verbose & 0.109 (0.008) & 0.098 (0.002) \\
 & express in speech & 0.833 (0.077) & 0.822 (0.076) \\
 & convert into a verb & 0.953 (0.037) & 0.102 (0.001) \\
 & articulate; either verbally or with a cry, shout, or noise & 0.975 (0.004) & 0.877 (0.003) \\
\midrule
\textbf{visualise} & view the outline of by means of an X-ray & 0.386 (0.038) & 0.225 (0.022) \\
 & make visible & 0.976 (0.080) & 0.734 (0.079) \\
 & imagine; conceive of; see in one's mind & 0.976 (0.003) & 0.700 (0.002) \\
 & form a mental picture of something that is invisible or abstract & 0.366 (0.309) & 0.162 (0.241) \\
\midrule
\textbf{voice} & the sound made by the vibration of vocal folds modified by the resonance of the vocal tract & 0.474 (0.014) & 0.282 (0.008) \\
 & an advocate who represents someone else's policy or purpose & 0.516 (0.002) & 0.308 (0.002) \\
 & expressing in coherent verbal form & 0.998 (0.003) & 0.889 (0.003) \\
 & the melody carried by a particular voice or instrument in polyphonic music & 0.992 (0.006) & 0.532 (0.004) \\
\midrule
\textbf{water} & supply with water, as with channels or ditches or streams & 0.009 (0.000) & 0.006 (0.000) \\
 & liquid excretory product & 0.980 (0.051) & 0.979 (0.051) \\
 & a facility that provides a source of water & 0.985 (0.048) & 0.784 (0.038) \\
 & the part of the earth's surface covered with water (such as a river or lake or ocean) & 0.990 (0.003) & 0.641 (0.002) \\
\midrule
\textbf{waxy} & having the paleness of wax; - Bram Stoker & 0.986 (0.001) & 0.083 (0.000) \\
 & easily impressed or influenced & 0.979 (0.046) & 0.896 (0.042) \\
 & made of or covered with wax & 0.012 (0.001) & 0.001 (0.000) \\
 & capable of being bent or flexed or twisted without breaking & 0.998 (0.003) & 0.803 (0.002) \\
\midrule
\textbf{weigh} & show consideration for; take into account & 0.936 (0.218) & 0.689 (0.087) \\
 & have weight; have import, carry weight & 0.646 (0.007) & 0.451 (0.005) \\
 & to be oppressive or burdensome; , & 0.992 (0.003) & 0.264 (0.000) \\
 & determine the weight of & 0.355 (0.000) & 0.245 (0.000) \\
\midrule
\textbf{work up} & come up with & 0.957 (0.017) & 0.938 (0.017) \\
 & form or accumulate steadily & 0.927 (0.041) & 0.043 (0.001) \\
 & bolster or strengthen & 0.091 (0.001) & 0.035 (0.000) \\
 & develop & 0.866 (0.246) & 0.605 (0.146) \\
\midrule
\end{longtable}

\normalsize

\end{document}